\documentclass[aip, jcp, twocolumn, floatfix, superscriptaddress, nofootinbib, reprint,10pt]{revtex4-1}

\newcommand{\bA}{\mathbf{A}}
\newcommand{\bAhat}{\hat{\mathbf{A}}}
\newcommand{\bAt}{\tilde{\mathbf{A}}}

\newcommand{\nf}{n_{\text{features}}}
\newcommand{\ns}{n_{\text{samples}}}
\newcommand{\np}{n_{\text{properties}}}
\newcommand{\npca}{n_{\text{latent}}}

\newcommand{\ntrain}{n_{\text{train}}}

\newcommand{\N}{N}
\newcommand{\M}{M}

\newcommand{\bP}[2]{\mathbf{P}_{{#1}{#2}}}
\newcommand{\Tt}{\tilde{T}}
\newcommand{\bPty}{\bP{T}{Y}}
\newcommand{\bPtty}{\bP{\Tt}{Y}}
\newcommand{\bPtx}{\bP{T}{X}}
\newcommand{\bPtk}{\bP{T}{K}}
\newcommand{\bPttx}{\bP{\Tt}{X}}
\newcommand{\bPxt}{\bP{X}{T}}
\newcommand{\bPxtt}{\bP{X}{\Tt}}
\newcommand{\bPgt}{\bP{K}{T}}

\newcommand{\bPxy}{\bP{X}{Y}}
\newcommand{\bPpy}{\bP{\Phi}{Y}}
\newcommand{\bPky}{\bP{K}{Y}}
\newcommand{\bPkt}{\bP{K}{T}}

\newcommand{\bX}{\mathbf{X}}

\newcommand{\bXprime}{\bX^{\prime}}
\newcommand{\bXbarprime}{\bar{\bX}^{\prime}}

\newcommand{\bx}{{\mathbf{x}}}

\newcommand{\bY}{\mathbf{Y}}
\newcommand{\bYprime}{\bY^{\prime}}
\newcommand{\bYbarprime}{\bar{\bY}^{\prime}}
\newcommand{\by}{\mathbf{y}}
\newcommand{\bYhat}{\hat{\bY}}

\newcommand{\bPhi}{\boldsymbol{\Phi}}
\newcommand{\bphi}{\boldsymbol{\phi}}

\newcommand{\bI}{\mathbf{I}}

\newcommand{\CX}{\mcal{X}}
\newcommand{\CA}{\mcal{A}}

\newcommand{\bC}{\mathbf{C}}
\newcommand{\bCt}{\tilde{\mathbf{C}}}

\newcommand{\bT}{\mathbf{T}}
\newcommand{\bt}{\mathbf{t}}
\newcommand{\bTt}{\tilde{\bT}}

\newcommand{\bG}{\mathbf{K}}
\newcommand{\bGt}{\tilde{\bG}}
\newcommand{\bGhat}{\hat{\bG}}

\newcommand{\bU}{\mathbf{U}}
\newcommand{\bUhat}{\hat{\bU}}

\newcommand{\bLAM}{\boldsymbol{\Lambda}}

\newcommand{\loss}[2]{\lVert {#1} - {#2} \rVert^2}
\newcommand{\lproj}{\loss{\bX}{\bT\bPtx}/\ns}
\newcommand{\lprojphi}{\loss{\bPhi}{\bT\bP{T}{\Phi}}/\ns}
\newcommand{\lregr}{\loss{\bY}{\bT\bPty}/\ns}
\newcommand{\lgram}{\loss{\bG}{\bT\bT^T}/\ns}

\newcommand{\Tr}{\operatorname{Tr}}

\newcommand{\mc}[1]{{ \color{blue}{ #1}}}
\newcommand{\MC}[1]{\mc{ \bf MC: #1}}

\newcommand{\RKC}[1]{{ \bf \color{red} {RKC: #1}}}

\newcommand{\bah}[1]{{\color{green}{BAH: #1}}}

\newcommand{\mcal}[1]{\ensuremath{\mathcal{#1}}}
\usepackage{bm}
\usepackage{graphicx}
\usepackage[usenames]{color}
\usepackage{microtype}
\usepackage{amsmath}
\usepackage{amssymb}
\usepackage{xspace}
\usepackage{comment}
\usepackage{footnote}

\usepackage[version=4]{mhchem}
\usepackage{soul}
\usepackage{relsize}
\usepackage{svg}
\usepackage{hhline}
\usepackage{verbatim}
\usepackage[normalem]{ulem}
\usepackage[inline]{enumitem}

\definecolor{mygray}{gray}{0.5}

\makeatletter
\g@addto@macro\normalsize{%
  \setlength\abovedisplayskip{5pt}
  \setlength\belowdisplayskip{5pt}
  \setlength\abovedisplayshortskip{5pt}
  \setlength\belowdisplayshortskip{5pt}
}
\makeatother
\usepackage{titlesec}

\titlespacing\section{0pt}{8pt plus 4pt minus 2pt}{2pt plus 2pt minus 2pt}
\titlespacing\subsection{0pt}{8pt plus 4pt minus 2pt}{2pt plus 2pt minus 2pt}
\titlespacing\subsubsection{0pt}{8pt plus 4pt minus 2pt}{2pt plus 2pt minus 2pt}

\usepackage{multirow}

\usepackage{array}

\makeatletter
\newcommand{\thickhline}{%
 \noalign {\ifnum 0=`}\fi \hrule height 1.5pt
 \futurelet \reserved@a \@xhline
}
\newcolumntype{"}{@{\hskip\tabcolsep\vrule width 1pt\hskip\tabcolsep}}
\makeatother

\begin{document}

\title{Structure-Property Maps with Kernel Principal Covariates Regression}

\author{Benjamin A. Helfrecht}
\affiliation{Laboratory of Computational Science and Modeling, IMX, \'Ecole Polytechnique F\'ed\'erale de Lausanne, 1015 Lausanne, Switzerland}

\author{Rose K. Cersonsky}
\affiliation{Laboratory of Computational Science and Modeling, IMX, \'Ecole Polytechnique F\'ed\'erale de Lausanne, 1015 Lausanne, Switzerland}

\author{Guillaume Fraux}
\affiliation{Laboratory of Computational Science and Modeling, IMX, \'Ecole Polytechnique F\'ed\'erale de Lausanne, 1015 Lausanne, Switzerland}

\author{Michele Ceriotti}
\email{michele.ceriotti@epfl.ch}
\affiliation{Laboratory of Computational Science and Modeling, IMX, \'Ecole Polytechnique F\'ed\'erale de Lausanne, 1015 Lausanne, Switzerland}
\begin{abstract}
Data analyses based on linear methods constitute the simplest, most robust, and transparent approaches to the automatic processing of large amounts of data for building supervised or unsupervised machine learning models.
Principal covariates regression (PCovR) is an underappreciated method that interpolates between principal component analysis and linear regression, and can be used to conveniently reveal structure-property relations in terms of simple-to-interpret, low-dimensional maps. 
Here we provide a pedagogic overview of these data analysis schemes, including the use of the kernel trick to introduce an element of non-linearity, while maintaining most of the convenience and the simplicity of linear approaches. 
We then introduce a kernelized version of PCovR and a sparsified extension, and demonstrate the performance of this approach in revealing and predicting structure-property relations in chemistry and materials science, showing a variety of examples including elemental carbon, porous silicate frameworks, organic molecules, amino acid conformers, and molecular materials.
\end{abstract}

\maketitle

\section{Introduction}
Over the past decade, there has been a tremendous increase in the use of data-driven and machine learning (ML) methods in materials science, ranging from the prediction of materials properties \cite{faber_machine_2016,faber_prediction_2017,hansen_assessment_2013,rupp_fast_2012}, to the construction of interatomic potentials \cite{deringer_machine_2017,dragoni_achieving_2018,maillet_machine-learning_2018,szlachta_accuracy_2014} and searches for new candidate materials for a particular application \cite{simon_materials_2015,sendek_holistic_2017,kahle_high-throughput_2020,kirklin_high-throughput_2013}.
Broadly speaking, these methods can be divided into two categories: those that are focused on predicting the properties of new materials (supervised learning), and those that are focused on finding or recognising patterns, particularly in atomic structures (unsupervised learning).
While supervised methods are useful for predicting properties of materials with diverse atomic configurations, they are not as well-suited for classifying structural diversity.
Conversely, unsupervised methods are useful for finding structural patterns, but often fail to directly predict materials properties.
Moreover, it can be difficult to validate motifs identified by an unsupervised learning algorithm, as the results obtained from the clustering algorithm depend on the choice of the structural representation and can therefore be biased by preconceived expectations on what the most relevant features should be~\cite{ceri19jcp}.

Methods that combine the predictive power of supervised ML and the pattern recognition capabilities of unsupervised ML stand to be very useful in materials informatics, making it possible to increase data efficiency and more clearly reveal structure-property relations.
A number of statistical methods have been developed for augmenting regression models to incorporate information about the structure of the input data, including principal component regression \cite{jolliffe_note_1982}, partial least squares regression \cite{wold_pls-regression:_2001}, cluster-wise regression \cite{spath_algorithm_1979}, continuum regression \cite{stone_continuum_1990}, and principal covariates regression (PCovR) \cite{de_jong_principal_1992,vervloet_selection_2013,vervloet_pcovr:_2015,vervloet_model_2016}.
Among these, PCovR is particularly appealing, because it transparently combines linear regression (LR; a supervised learning method) with principal component analysis (PCA; an unsupervised learning method). The method has found previous applications in climate science \cite{fischer_regularized_2014}, macroeconomics \cite{HEIJ20073612}, social science \cite{vervloet_pcovr:_2015}, and bio-informatics \cite{van_deun_obtaining_2018, taylor_sullivan_ellerbeck_gajewski_gibbs_2019}, but has yet to be widely adopted. A handful of extensions have been developed for PCovR, including a combination with cluster-wise regression \cite{wilderjans_principal_2017}, and regularised models \cite{fischer_regularized_2014,van_deun_obtaining_2018}.

In this paper, we propose a kernel-based variation on the original PCovR method, which we call \textit{Kernel Principal Covariates Regression} (KPCovR), with the aim of making it even more versatile for statistics and machine learning applications.
We begin by summarising the required background concepts and constituent methods used in the construction of linear PCovR in addition to the kernel trick, which can be used to incorporate an element of non-linearity in otherwise linear methods. We then introduce KPCovR, both for full and sparse kernels, and demonstrate their application to several different classes of materials and chemical systems.

\section{Background Methods}
\label{sec:methods}
We start by giving a concise but complete overview of established linear methods for dimensionality reduction and regression, as well as their kernelized counterparts. This is done to set a common notation and serve as a pedagogic introduction to the problem, complemented by a set of interactive Jupyter notebooks~\cite{kpcovr-notebooks}.
Expert readers can skip this section and proceed to Section~\ref{sec:extensions}, where we introduce kernelized PCovR methods.
Throughout this section, we demonstrate the methods on the CSD-1000r dataset \cite{CSD1000}, which contains the NMR chemical shielding of nuclei in a collection of 1000 organic crystals and their 129,580 atomic environments, of which we use 25,600 in this study. To simplify this into a more illustrative example, we classify and predict simultaneously the chemical shieldings of all nuclei, even though in actual applications one usually would deal with one element at a time.
As the input features, we use the SOAP power spectrum vectors, which discretise a three-body correlation function including information on each atom, its relationships with neighbouring atoms, and the relationships between sets of neighbours \cite{Bartok2013, will+19jcp}.

\begin{figure}[tbhp]
    \centering
    \includegraphics[width=\linewidth]{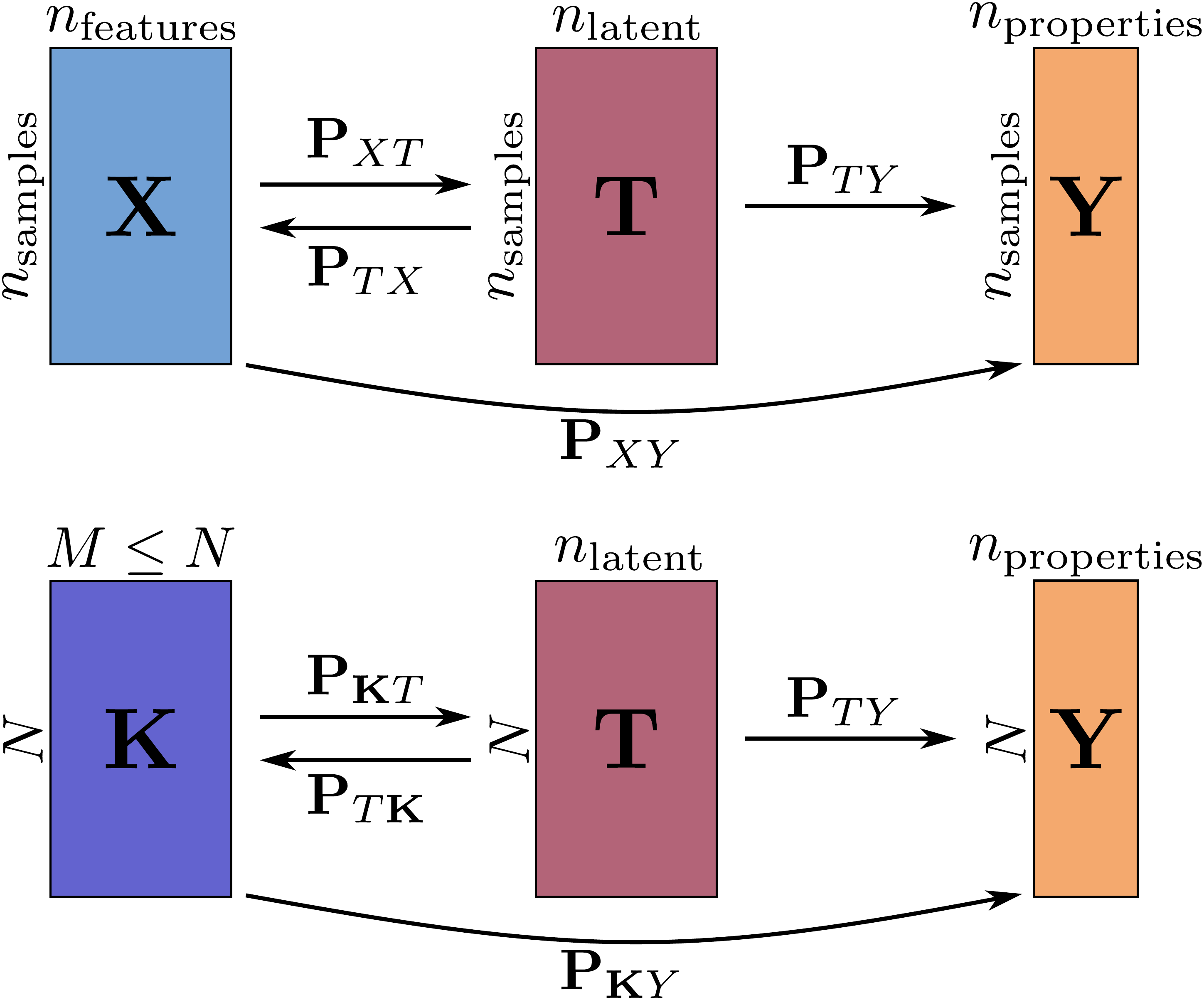}
    \caption{A schematic representation of the different linear operations that can be performed to model structure-property relations in terms of a matrix of features $\bX$ that represents the input samples.}
    \label{fig:matrices}
\end{figure}

\subsection{Notation}

In the following, we assume that the input data has been processed in such a way that the nature of each sample (e.g., the composition and structure of a molecule) is encoded as a row of a feature matrix $\bX$.
Each sample is therefore a vector $\bx$ of length $\nf$, so that $\bX$ has the shape $\ns \times \nf$. Similarly, the properties associated with each sample are stored in a property matrix $\bY$, which has the shape $\ns \times \np$.
We denote the data in latent space (i.e., a low-dimensional approximation of $\bX$) as $\bT$. We denote each projection matrix from one space to another as $\bP{A}{B}$, where $A$ is the original space and $B$ is the projected space. As such, the projection matrix from the input space to $\bT$ is $\bPxt$, and vice versa $\bPtx$ is the projection matrix from $\bT$ to the input space. Note that in general projectors $\bP{A}{B}$ are not assumed to be orthogonal nor full-rank. A graphical summary of the mappings that we consider in this paper is depicted in Figure~\ref{fig:matrices}.

To simplify notation and to work with unit-less quantities, we assume in our derivations that both $\bX$ and $\bY$ are centred according to their respective column means and are scaled to have unit variance.
A similar centring and scaling procedure is also applied when working with kernels~\cite{scholkopf_nonlinear_1998}. Centring and scaling is discussed in more detail in appendix~\ref{app:centring}, and demonstrated in the companion Jupyter notebooks~\cite{kpcovr-notebooks}.
To make notation less cumbersome, variables names are not defined uniquely across the entirety of the paper. %
We re-use variable names for common elements among the different subsections---for example, using $\bT$ to represent a low-dimensional latent space in all methods---but the precise definitions of the re-used variables may differ between subsections and should not be confused with one another.

We also use throughout a few additional conventions: (1) we write an approximation or truncation of a given matrix $\bA$ as $\bAhat$; (2) we use $\bAt$ to signify an augmented version of $\bA$; that is, $\bAt$ is defined differently from $\bA$, but occupies the same conceptual niche (3)
we represent the eigendecomposition of a symmetric matrix as $\bA = \bU_{\bA} \bLAM_{\bA} \bU_{\bA}^T$,  where $\bLAM_{\bA}$ is a diagonal matrix containing the eigenvalues and $\bU_{\bA}$ the matrix having the corresponding eigenvectors as columns; (4) we use throughout the Frobenius norm $\lVert\bA\rVert=\sqrt{\Tr\bA^T\bA}$; and (5) we define and report the values of the different losses normalised by the number of sample points, but we omit such normalisation in derivations to declutter equations.

\subsection{Linear Methods}
\label{sec:linear}
We begin by discussing models of the form:
\begin{equation}
    \mathbf{B} = \mathbf{A}\bP{A}{B}
\end{equation}
where $\mathbf{B}$ is a target quantity (e.g., a property that one wants to predict or an alternative, lower-dimensional representation of the feature matrix $\mathbf{A}$), and $\bP{A}{B}$ is a linear projection that maps the features to the target quantity\cite{esl, bishop}.

\begin{figure}[tbph]
    \centering
    \includegraphics[width=\linewidth]{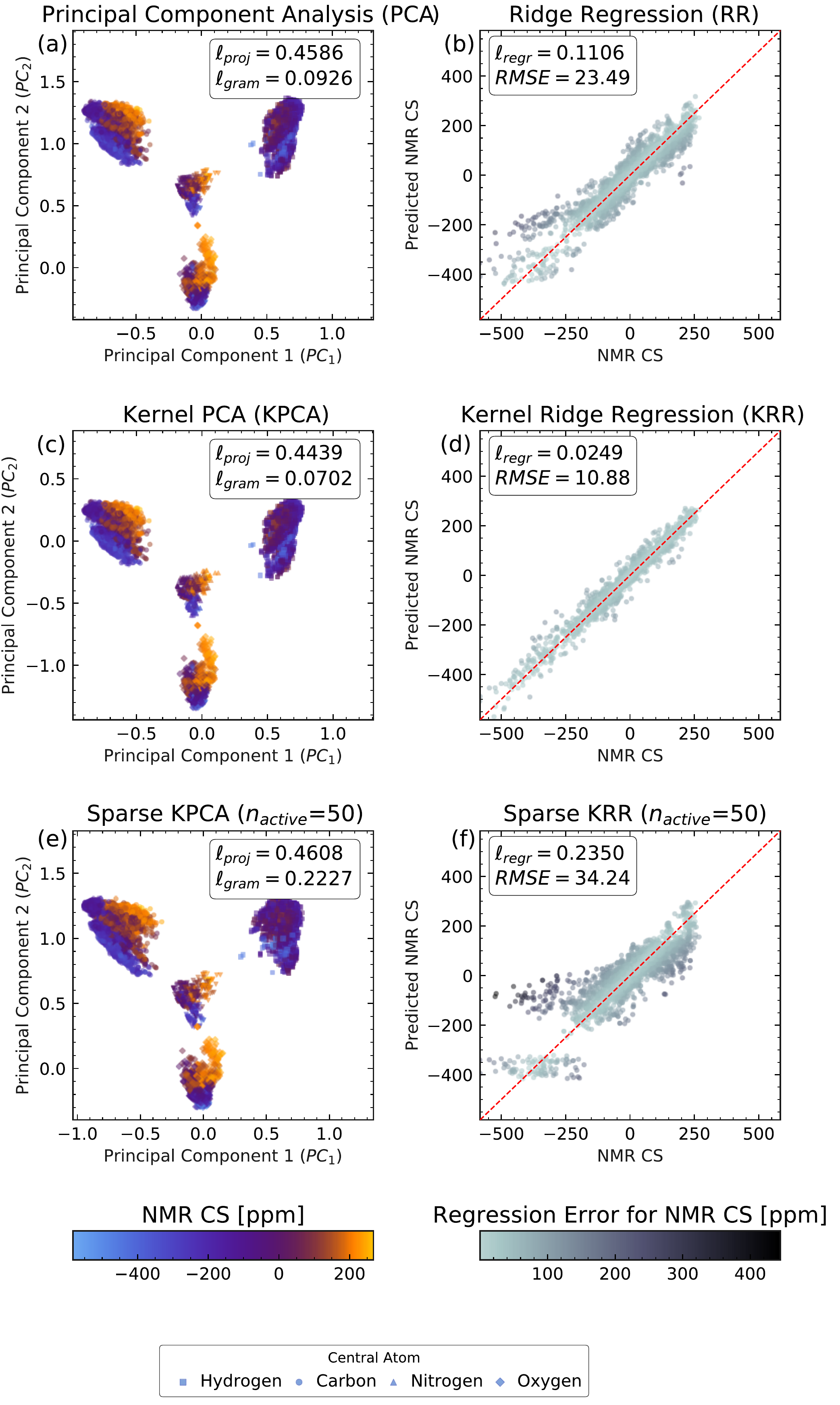}
    \caption{\textbf{Projection and Regression Models of CSD-1000r.} In each projection, the property values are denoted by marker colour and the marker symbol denotes the central atom of each environment, which corresponds to the cluster in the projection. In each regression, the target is denoted by a dotted line. colours denote absolute error of predicted properties, and inset includes the values of $\ell_\text{proj} = \lproj$, $\ell_\text{regr} = \lregr$, and root-mean-square error (RMSE), where appropriate. Projections: (a) Principal Components Analysis (PCA) and Multidimensional Scaling (MDS), (c) Kernel PCA, and (e) Sparse Kernel PCA, with $n_{active}=50$.  Regressions: (b) Ridge Regression, (d) Kernel Ridge Regression (KRR), and (f) Sparse KRR, with $n_{active}=50$. 
    }
    \label{fig:models}
\end{figure}

\subsubsection{Principal Component Analysis}
\label{sec:pca}

In principal component analysis~\cite{f.r.s_liii._1901,hotelling_analysis_1933}, the aim is to reduce the dimensionality of the feature matrix $\bX$ by determining the orthogonal projection $\bT = \bX\bPxt$ which incurs minimal information loss. More formally, we wish to minimise the error $\ell$ of reconstructing $\bX$ from the low-dimensional projection:
\begin{equation}
    \ell_\text{proj} = \lproj.
    \label{eq:loss-pca}
\end{equation}
The requirement that $\bPxt$ is orthonormal implies that $\bPtx = \bPxt^T$.
Using the properties of the Frobenius norm, $\ell$ can be rewritten as
\begin{equation}
    \ell = \Tr\left(\bX\left(\bI - \bPxt\bPxt^T\right)\bX^T\right)
\end{equation}
which is minimised when the similarity
\begin{equation}
    \rho = \Tr(\bPxt^T\bX^T\bX\bPxt)
\end{equation}
is maximised. Given the orthogonality constraint on $\bPxt$, the similarity is maximised when $\bPxt$ corresponds to the eigenvectors of the covariance $\bC = \bX^T\bX$ that are associated with the $\npca$ largest eigenvalues. We introduce the eigendecomposition $\bC = \bU_\bC\bLAM_\bC\bU_\bC^T$, where $\bU_\bC$ is the matrix of the eigenvectors and $\bLAM_\bC$ the diagonal matrix of the eigenvalues, so that
\begin{equation}
    \bT = \bX\bUhat_\bC,
\label{eq:txuc}
\end{equation}
where we use the notation $\bUhat$ to indicate the matrix containing only the top $\npca$ components.
The outcomes of a PCA with $\npca=2$ of the CSD-1000r dataset are shown in Fig.~\ref{fig:models}(a).
The atomic environments are split clearly according to the nature of the atom sitting at the centre of the environment, reflecting the prominence of this information in the SOAP features we use.

\subsubsection{Multidimensional scaling}
\label{sec:mds}

A reduction in the dimensionality of the feature space can also be achieved with a different logic that underlies several methods grouped under the label of multidimensional scaling (MDS) \cite{Torgerson1952}.
In MDS, the projected feature space is chosen to preserve the pairwise distances of the original space, defining the loss
\begin{equation}
    \ell = \frac{1}{\ns}\sum_{i<j} \left( \left|\bx_i - \bx_j\right|^2 - \left|\bt_i - \bt_j\right|^2 \right)^2,
\label{eq:loss-mds-sum}
\end{equation}
where $\bx_i$ and $\bt_i$ refer to the full and projected feature vector of the $i$-th sample.
In general, Eq.~\eqref{eq:loss-mds-sum} requires an iterative optimisation. When the distance between features is the Euclidean distance, as in classical MDS, the link between the metric and the scalar product suggests minimising the alternative loss
\begin{equation}
    \ell_\text{gram} = \lgram
\label{eq:loss-mds}
\end{equation}
Note that the solutions of Eqs.~\eqref{eq:loss-mds-sum} and~\eqref{eq:loss-mds} concur only if one can find a solution that zeroes $\ell$.
If the eigenvalue decomposition of the Gram matrix reads $\bG = \bX \bX^T = \bU_\bG \bLAM_\bG \bU_\bG^T$, $\ell$ is minimised when $\bT\bT^T$ is given by the singular value decomposition of $\bG$, that is by taking
\begin{equation}
\bT = \hat{\bU}_\bG \hat{\bLAM}_\bG^{1/2}
\label{eq:tuk}
\end{equation}
restricted to the largest $\npca$ eigenvectors.
However, $\bC$ and $\bG$ have the same (non-zero) eigenvalues, and the (normalised) eigenvectors are linked by $\bU_\bG = \bX\bU_\bC\bLAM_\bC^{-1/2}$.
Hence, one sees that $\bT = \bX \hat{\bU}_\bC$, consistent with Eq.~\eqref{eq:txuc}. Thus, classical MDS  yields the same result as PCA in Fig.~\ref{fig:models}(a).

\subsubsection{Linear Regression}
\label{sec:lr}
In linear regression, one aims to determine a set of weights $\bPxy$ to minimise the error between the true properties $\bY$ and the properties predicted via $\bYhat = \bX\bPxy$, which is equivalent to minimising the loss
\begin{equation}
    \ell_\text{regr} =\lregr
    \label{eq:loss-lr}
\end{equation}
In the following, we consider the case of an $\mathcal{L}^2$ regularised regression with regularisation parameter $\lambda$, i.e., ridge regression \cite{esl}. The loss to be minimised is
\begin{equation}
    \ell = \loss{\bY}{\bX\bPxy} + \lambda\lVert\bPxy\rVert^2.
\end{equation}

Minimising the loss with respect to $\bPxy$ yields the solution $\bPxy = \left(\bX^T\bX+\lambda\bI\right)^{-1}\bX^T\bY$.
If one chooses to perform the regression using the low-dimensional latent space $\bT=\bX \bPxt$ and approximate $\bY$ with $\bT\bPty$, then $\bPty = \left(\bT^T\bT+\lambda\bI\right)^{-1}\bT^T\bY$.

The ridge regression of the CSD-1000r dataset is shown in Fig.~\ref{fig:models}(b). Given the small train set size, and the difficulty of fitting simultaneously  different elements with shieldings across a large range ($\approx$ 800 ppm), the model achieves a very good accuracy, with a RMSE below 23 ppm.

\begin{figure*}[tbph]
    \centering
    \includegraphics[width=\linewidth]{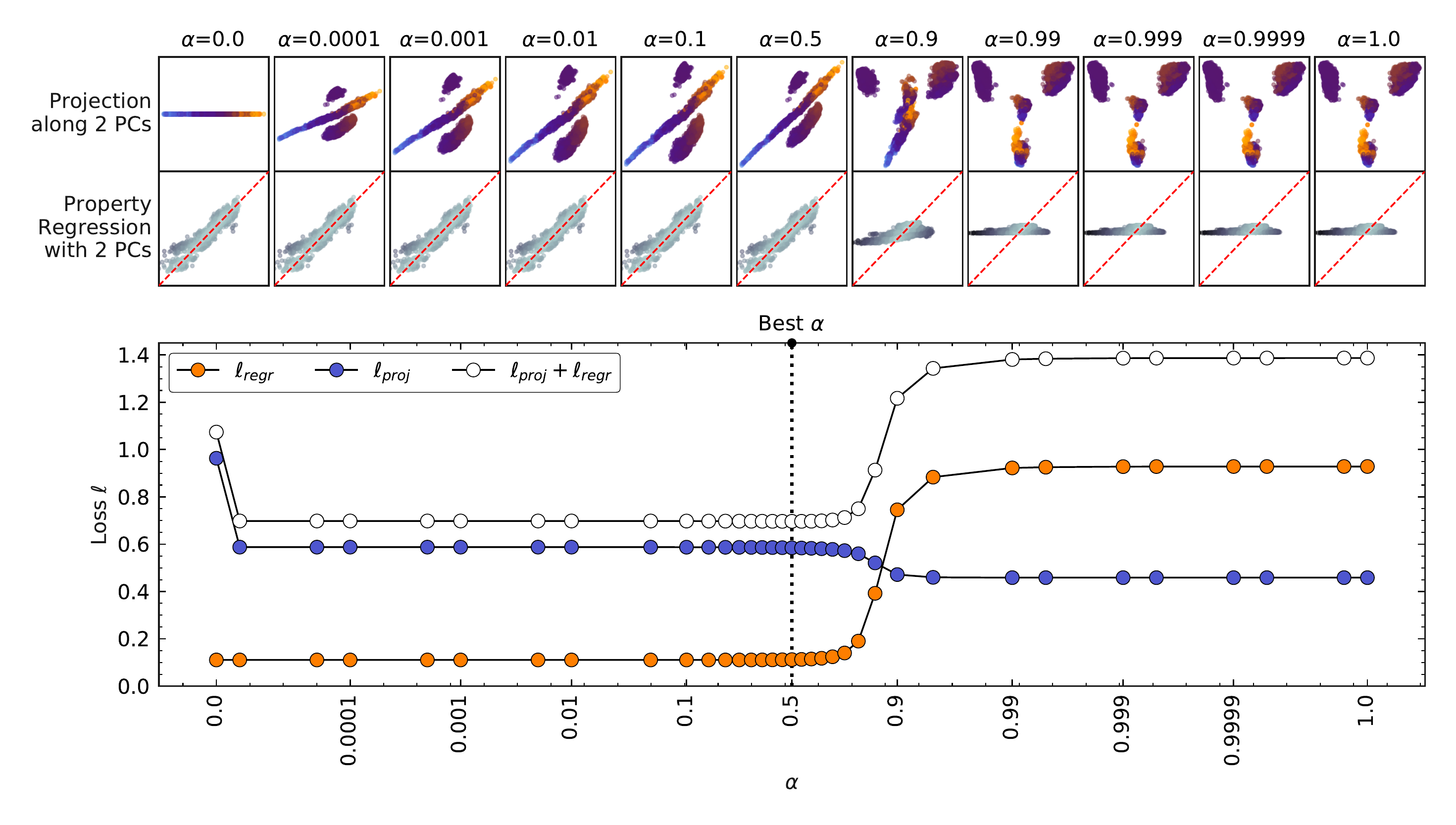}
    \caption{\textbf{Principal Covariates Regression of CSD-1000r.} Combining ridge regression (far left) and PCA (far right) with mixing parameter $\alpha$, PCovR can minimise the total loss $\ell = \ell_\text{proj} +\ell_\text{regr} = \lproj + \lregr$, as denoted in white in the figure. The upper panels show the resulting projections and regressions at the indicated $\alpha$ value, aligned with the horizontal axis in the lower plot. Colour mappings correspond to those in Fig.~\ref{fig:models} and Fig.~\ref{fig:kpcovr-models}. An animated representation of the top panels can be found in the SI.}
    \label{fig:PCovR}
\end{figure*}

\subsection{Principal Covariates Regression}
\label{sec:pcovr}

Principal covariates regression (PCovR)~\cite{de_jong_principal_1992} utilises a combination between a PCA-like and a LR-like loss, and therefore attempts to find a low-dimensional projection of the feature vectors that simultaneously minimises information loss and error in predicting the target properties using only the latent space vectors $\bT$.
A mixing parameter $\alpha$ determines the relative weight given to the PCA and LR tasks,
\begin{equation}
    \ell = \frac{\alpha}{\ns}\loss{\bX}{\bX\bPxt\bPtx} + \frac{(1 - \alpha)}{\ns}\loss{\bY}{\bX\bPxt\bPty}.
    \label{eq:loss-pcovr}
\end{equation}
The derivation we report here, albeit in our notation, follows closely that in the original article~\cite{de_jong_principal_1992}. PCovR can be implemented in a way that diagonalises a modified Gram matrix (sample-space PCovR) or in a way that requires computing and diagonalising a modified covariance (feature-space PCovR). The two approaches yield the same latent-space projections, and which one should be used depends on the relative magnitudes of $\ns$ and $\nf$.

\subsubsection{Sample-space PCovR}

It is easier to minimise Eq.~\eqref{eq:loss-pcovr} by looking for a projection $\bTt$ in an auxiliary latent space for which we enforce orthonormality, $\bTt^T\bTt = \bI$, known as a \textit{whitened} projection.

This allows us to write $\bPttx = \bTt^T\bX$ and $\bPtty = \bTt^T\bY$. By definition $\bTt = \bX\bPxtt$, thus we can express the loss as
\begin{equation}
    \ell = \alpha\loss{\bX}{\bTt\bTt^T\bX} + (1 - \alpha)\loss{\bY}{\bTt\bTt^T\bY}.
\end{equation}
This loss is minimised by maximising the associated similarity
\begin{align}
    \rho &= \Tr\left(\alpha\bTt\bTt^T\bX\bX^T + (1 - \alpha)\bTt\bTt^T\bYhat\bYhat^T\right)\\
         &= \Tr\left(\alpha\bTt\bTt^T\bX\bX^T + (1 - \alpha)\bTt\bTt^T\bX\bPxy\bPxy^T\bX^T\right),
\end{align}
where we have substituted $\bY$ with the regression approximation $\bYhat=\bX\bPxy$ --- given that a linear approximation of $\bY$ in the latent space can only, at best, reproduce the part of the properties that can be represented in the full feature space.
If we define the modified Gram matrix
\begin{equation}
\bGt = \alpha\bX\bX^T + (1 - \alpha)\bX\bPxy\bPxy^T\bX^T,
\label{eq:gram-pcovr}
\end{equation}
we can further write the similarity as

\begin{equation}
    \rho = \Tr\left(\bTt^T\bGt\bTt\right).
\label{eq:rho-pcovr}
\end{equation}

The latent space projections $\bTt$ that maximise the similarity correspond to the principal eigenvectors of the matrix $\bGt$, $\bTt=\hat{\bU}_{\bGt}$.
By analogy with multidimensional scaling---and to ensure that in the limit of $\alpha\rightarrow 1$ we obtain the same latent space as in classical MDS---one can obtain de-whitened projections $\bT = \hat{\bU}_{\bGt} \hat{\bLAM}_{\bGt}^{1/2}=\bGt \hat{\bU}_{\bGt} \hat{\bLAM}_{\bGt}^{-1/2}$, reminiscient to Eq.~\eqref{eq:tuk}.
The projector from feature space to the latent space is then given by
\begin{equation}
    \bPxt = \left(\alpha\bX^T + (1 - \alpha)\bPxy\bPxy^T\bX^T\right)\hat{\bU}_{\bGt}\hat{\bLAM}_{\bGt}^{-1/2}.
    \label{eq:Pxt}
\end{equation}

The projector matrix from the latent space to the properties $\bY$ can be computed from the LR solution
\begin{equation}
    \bPty = \left(\bT^T\bT+\lambda\bI\right)^{-1}\bT^T\bY \underset{\lambda\rightarrow 0}{=} \hat{\bLAM}_{\bGt}^{-1/2} \hat{\bU}_{\bGt}^T\bY.
    \label{eq:Pty}
\end{equation}

\subsubsection{Feature-space PCovR}

Rather than determining the optimal PCovR projections by diagonalising the equivalent of a Gram matrix, one can tackle the problem in a way that more closely resembles PCA by instead diagonalising a modified covariance matrix.
Given that  $\bI = \bTt^T\bTt = \bPxtt^T\bX^T\bX\bPxtt = \bPxtt^T\bC\bPxtt$, we see that $\bC^{1/2}\bPxtt$ is orthogonal.
We can thus rewrite the similarity function from Eq.~\eqref{eq:rho-pcovr} as
\begin{equation}
    \rho = \Tr\left(\bPxtt^T\bC^{1/2}\bCt \bC^{1/2}\bPxtt\right),
\end{equation}
introducing
\begin{equation}
    \begin{split}
    \bCt = & \bC^{-1/2}\bX^T\bGt\bX\bC^{-1/2}  \\
    =&\alpha\bC + (1 - \alpha)\bC^{-1/2}\bX^T\bYhat\bYhat^T\bX\bC^{-1/2} \\
    =&\bU_{\bCt}\bLAM_{\bCt}\bU_{\bCt}^T.
    \end{split}
    \label{eq:pcovr-covariance}
\end{equation}
The similarity is maximised when the orthogonal matrix $\bC^{1/2}\bPxtt$ matches the principal eigenvalues of $\bCt$, i.e. $\bPxtt = \bC^{-1/2}\hat{\bU}_{\bCt}$.
In general
$\bPxtt \bPttx = \bC^{-1/2}\hat{\bU}_{\bCt}\hat{\bU}_{\bCt}^T \bC^{1/2}$ is not a symmetric matrix, and so it is not possible to define an orthonormal $\bPxt$ such that $\bPtx=\bPxt^{T}$.
Consistently with the case of sample-space PCovR, we obtain
\begin{equation}
    \begin{split}
    \bPxt =& \bC^{-1/2}\hat{\bU }_{\bCt}\hat{\bLAM}_{\bCt}^{1/2} \\
    \bPtx =& \hat{\bLAM}_{\bCt}^{-1/2} \hat{\bU}_{\bCt}^T  \bC^{1/2} \\
    \bPty =& \hat{\bLAM}_{\bCt}^{-1/2} \hat{\bU}_{\bCt}^T  \bC^{-1/2} \bX^T \bY, \\
    \end{split}
    \label{eq:pcovr-projectors}
\end{equation}
which minimise the PCovR loss in Eq.~\eqref{eq:loss-pcovr}. These projections reduce to PCA as $\alpha\rightarrow 1 $ and---if the dimension of the latent space is at least as large as the number of target properties in $\bY$---reduce to LR as $\alpha\rightarrow 0 $.

Figure~\ref{fig:PCovR} demonstrates the behaviour of PCovR when applied to the analysis of the CSD-1000r dataset. Here we plot $\ell_\text{proj}$ and $\ell_\text{regr}$ as a function of $\alpha$. The thumbnails above the losses correspond to the projections $\bT = \bX \bPxy$ and regressions $\bYhat = \bT \bPty$ for the indicated $\alpha$ below. Animations of these thumbnails are also given in the SI in \textit{gif} format.

For $\alpha=0$, we recover the accuracy of pure LR in predicting the values of the chemical shielding, but obtain a latent space that misses completely the structure of the dataset. The first principal component reflects the LR weight vector $\bPxy$, and the second carries no meaningful information.
For $\alpha=1$, we recover the PCA projection, that separates clearly the environments based on the nature of the central atom. A linear model built in the two-dimensional latent space, however, performs very poorly, because there is no \emph{linear} correlation between the position in latent space and the shielding values.
Intermediate values of $\alpha$ yield a projection that achieves the best of both worlds. The regression error is close to that of pure LR, but the error in the reconstruction of the input data from the latent space is now only marginally increased compared to pure PCA.

The PCovR map that corresponds to this ``optimal'' value of $\alpha$ achieves $\ell_\text{proj} = 0.585$ (comparable to the PCA value of $0.460$) and $\ell_\text{proj} = 0.112$ (comparable to the LR value of of $0.111$). Considering the poor performance of PCA in regression ($\ell_\text{regr} = 0.928$) and LR in projection ($\ell_\text{proj} = 0.963$), it is clear that the latent-space description of the dataset achieves a more versatile representation of structure-property relations.
There is still a recognisable clustering of the environments according to central atom species, but the O cluster, that exhibits the largest variance in the values of the shielding, is spread out diagonally so as to achieve maximal correlation between the position in latent space and value of the target properties.
We propose that -- in the absence of specific reasons suggesting to emphasise solely the regression or the projection accuracy -- an optimal value of $\alpha$ can be obtained looking for the minimum in $\ell_\text{proj}+\ell_\text{regr}$.

\subsection{Kernel Methods}
\label{sec:kernel}

While linear methods have the beauty of simplicity, they rely on the knowledge of a sufficient number of informative features that reflect the relation between inputs and properties.  Kernel methods introduce a possibly non-linear relation between samples in the form of a positive-definite kernel function $k(\bx, \bx ')$ (e.g. the Gaussian kernel $\exp(-\loss{\bx}{\bx'})$, or the linear kernel $\bx\cdot \bx'$), and use it to define a higher-dimensional space in which data points serve effectively as an adaptive basis~\cite{cutu10arxiv}. Unless otherwise specified, here we use a radial basis function (RBF) kernel, $\exp(-\gamma\loss{\bx}{\bx'})$, with the hyperparameter $\gamma$ optimized for each data set, as shown in the SI.
Doing so can help uncover non-linear relationships between the samples, resulting ultimately in more effective determination of a low-dimensional latent space and increased regression performance.

Mercer's theorem \cite{mercer_xvi._1909} guarantees that given a positive definite kernel there is a linear operator $\bphi(\bx)$ that maps input features into a (possibly infinite-dimensional) reproducing kernel Hilbert space (RKHS)~\cite{cutu10arxiv} whose scalar product generates the kernel, i.e. $\bphi(\bx)\cdot \bphi(\bx') = k(\bx,\bx')$.
$\bphi(\bx)$ is not necessarily known explicitly, but as we will see it can be approximated effectively for a given dataset, and we will use the notation $\bPhi$ to indicate the feature matrix that contains the (approximate) values of the kernel features for all of the sample points.
We indicate with $\bG=\bPhi\bPhi^T$ the $\ns\times\ns$ matrix that contains as entries the values of the kernel function between every pair of samples. In the case of a linear kernel, this is simply the Gram matrix computed for the input features, while for a non-linear kernel its entries can be computed by evaluating the kernel between pairs of samples, $K_{ij}=k(\bx_i,\bx_j)$.
Analogously to what we did for linear methods, we centre and normalise all the kernels we use in this work. Some subtleties connected to the centring operation are discussed in Appendix~\ref{app:centring}

\subsubsection{Kernel Principal Component Analysis}
\label{sec:kpca}

Kernel principal component analysis~\cite{scholkopf_nonlinear_1998} proceeds parallel to classical MDS, to which it corresponds exactly when a linear kernel is used. To construct a KPCA decomposition, one computes the eigendecomposition of the kernel matrix $\bG=\bU_\bG \bLAM_\bG \bU_\bG^T$ and defines the projections as the principal components $\bT=\hat{\bU}_\bG \hat{\bLAM}_\bG^{1/2}$. \footnote{If one retains all the $\ns$ eigenvectors, $\bT$ corresponds to an exact approximation of the kernel features for the given dataset, as $\bT\bT^T=\bG=\bPhi\bPhi^T$.}
The projections can also be computed as $\bT=\bG\bPgt=\bG\hat{\bU}_\bG \hat{\bLAM}_\bG^{-1/2}$, and this second expression can be used to project new data (in place of $\bG$ we use the matrix containing the values of the kernel matrix between new and reference points) in the approximate RKHS defined by the original samples. One can also approximate the kernel using the projector $\bPtk = \hat{\bLAM}_\bG^{-1}\bT^T \bG$.
As shown in Fig.~\ref{fig:models}c, for this dataset there is little qualitative difference between what we obtained with plain PCA and the KPCA projection. This is because SOAP features exhibit very clear correlations with the nature of the central environments, which is already well represented with a linear model.  While it is possible to compute the loss  $\ell_\text{proj} = \lproj$ associated with the approximation of $\bX$ based on $\bT$, the aim of KPCA is to approximate the kernel, and it is more appropriate to judge the methods performance based on a Gram loss $\ell_\text{gram}=\lgram$, which reduces to Eq.~\eqref{eq:loss-mds} for linear kernels.
Alternatively, one can compute a projection loss based on the approximation of RHKS features,
$\ell_\text{proj} = \lVert \bPhi  - \bT {\bP{T}{\Phi}} \rVert^2/\ns$, as detailed in Appendix~\ref{app:kpcovr-loss}.

\subsubsection{Kernel Ridge Regression}
\label{sec:krr}

Kernel ridge regression \cite{girosi_regularization_1995,smola_sparse_2000} is analogous to ridge regression, except that the kernel feature space vectors $\bPhi$ are substituted for the original input data $\bX$, giving the loss
\begin{equation}
\ell = \loss{\bY}{\bPhi\bPpy} + \lambda\lVert\bPpy\rVert^2,
\label{eq:krr-loss}
\end{equation}
so that the optimal weights are
\begin{equation}
    \begin{split}
    \bPpy &= \left(\bPhi^T\bPhi + \lambda \bI\right)^{-1}\bPhi^T\bY \\
    &= \bPhi^T\left(\bPhi\bPhi^T + \lambda\bI\right)^{-1}\bY.
    \end{split}
\end{equation}
Predicted properties $\bYhat$ can then be evaluated with $\bYhat = \bPhi\bPpy$. One can avoid computing explicitly the RKHS features by redefining the weights as $\bPky = \left(\bPhi\bPhi^T + \lambda\bI\right)^{-1}\bY = \left(\bG + \lambda\bI\right)^{-1}\bY$ so that $\bPpy = \bPhi^T\bPky$.\cite{mlpp} We can then write the predicted properties as
\begin{equation}
    \bYhat = \bPhi\bPhi^T\bPky = \bG\bPky.
\end{equation}
As shown in Fig.~\ref{fig:models}d, the greater flexibility afforded by a kernel model reduces the error by over 70\%{}.

\subsection{Sparse Kernel Methods}
\label{sec:sparse}

Since the size of kernel matrices grows in $n^2$ with respect to the number of samples, one wants to avoid computing (and inverting) the whole kernel matrix for large datasets. Instead, we can formulate a low-rank approximation to the kernel matrix through the Nystr\"om approximation \cite{williams_using_2001}, using a sub-selection of the data points, the active set, to define an approximate RKHS. These representative points can be selected in a variety of ways; two straightforward methods that have been used successfully in atomistic modelling are farthest point sampling (FPS) \cite{FPS} and a CUR matrix decomposition \cite{Mahoney697,bart-csan15ijqc, imba+18jcp}.

Using the subscript $\N$ to represent the full set of training data and $\M$ to indicate the active set, one can explicitly construct the approximate feature matrix as $\bPhi_{\N\M} = \bG_{\N\M}{\bU}_{\bG_{\M\M}}{\bLAM}_{\bG_{\M\M}}^{-1/2}  $,  where $\bU_{\M\M}$ and $\bLAM_{\M\M}$ are from the eigendecomposition of $\bG_{\M\M}$.
All sparse kernel methods can be derived in terms of a linear method based on the RKHS, although it is often possible to avoid explicitly computing $\bPhi_{\N\M}$. For instance, the approximate kernel matrix takes the form \cite{williams_using_2001}
\begin{equation}
    \bG \approx \bGhat_{\N\N} = \bPhi_{\N\M}\bPhi_{\N\M}^T =  \bG_{\N\M}\bG_{\M\M}^{-1}\bG_{\N\M}^T.
\end{equation}
For the following methods, we consider the approximate feature matrix $\bPhi_{\N\M}$ to be centred and scaled as discussed in Appendix~\ref{app:centring}).

\begin{figure*}[tbp]
    \centering
    \includegraphics[width=1.0\linewidth]{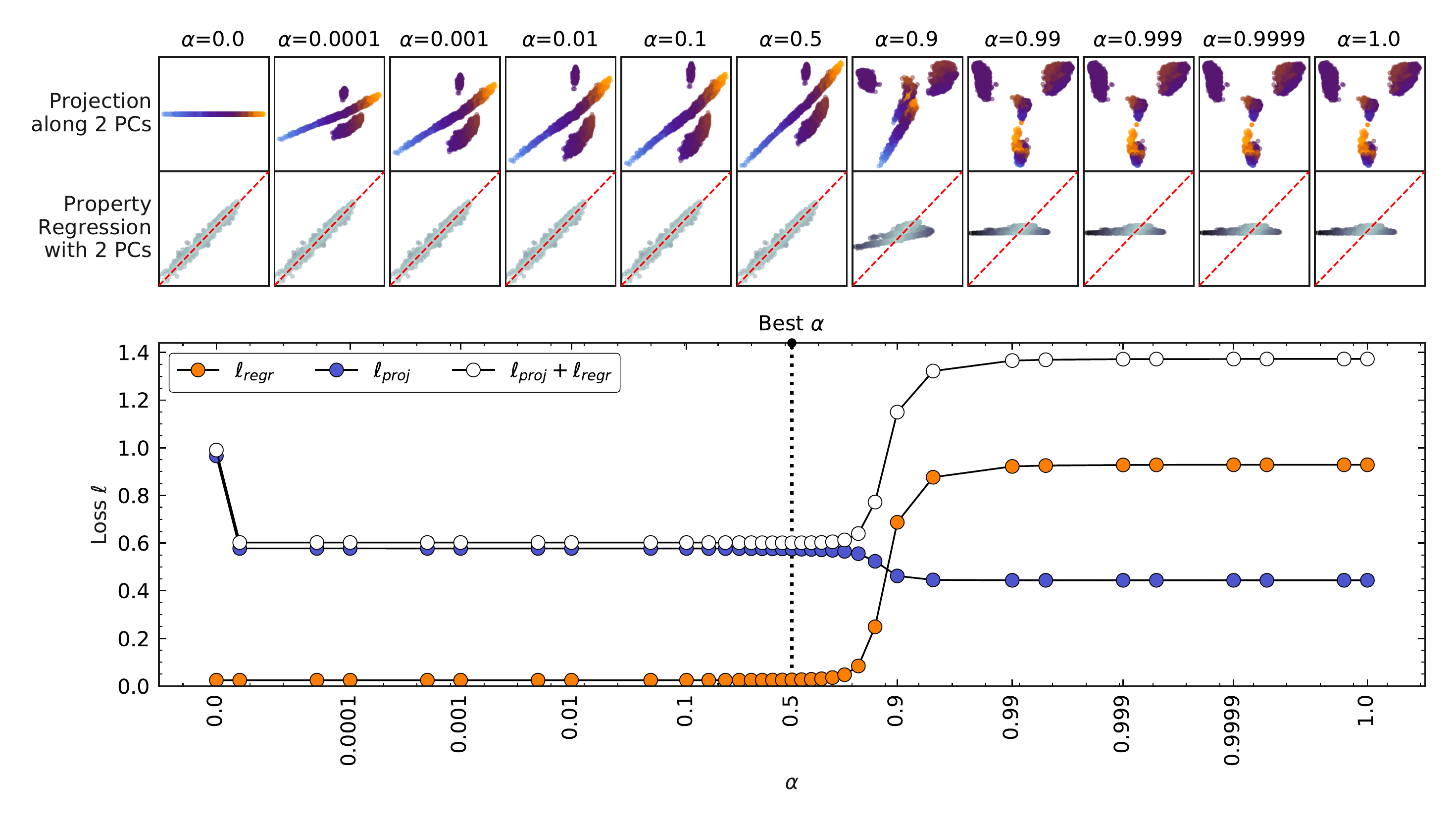}
    \caption{\textbf{Kernel Principal Covariates Regression of CSD-1000r.} Combining kernel ridge regression (far left) and KPCA (far right) with mixing parameter $\alpha$. Combined loss $\ell = \ell_\text{proj} + \ell_\text{regr} = \lprojphi + \lregr$ is given in the lower panel. Similar to Fig.~\ref{fig:PCovR}, the upper panels show the resulting projections and regressions at the indicated $\alpha$ value, aligned with the horizontal axis in the lower plot. colour mappings again correspond to those in Fig.~\ref{fig:models} and Fig.~\ref{fig:kpcovr-models}. An animated representation of the top panels can be found in the SI.}
    \label{fig:KPCovR}
\end{figure*}

\begin{figure}[tbp]
    \centering
    \includegraphics[width=1.0\linewidth]{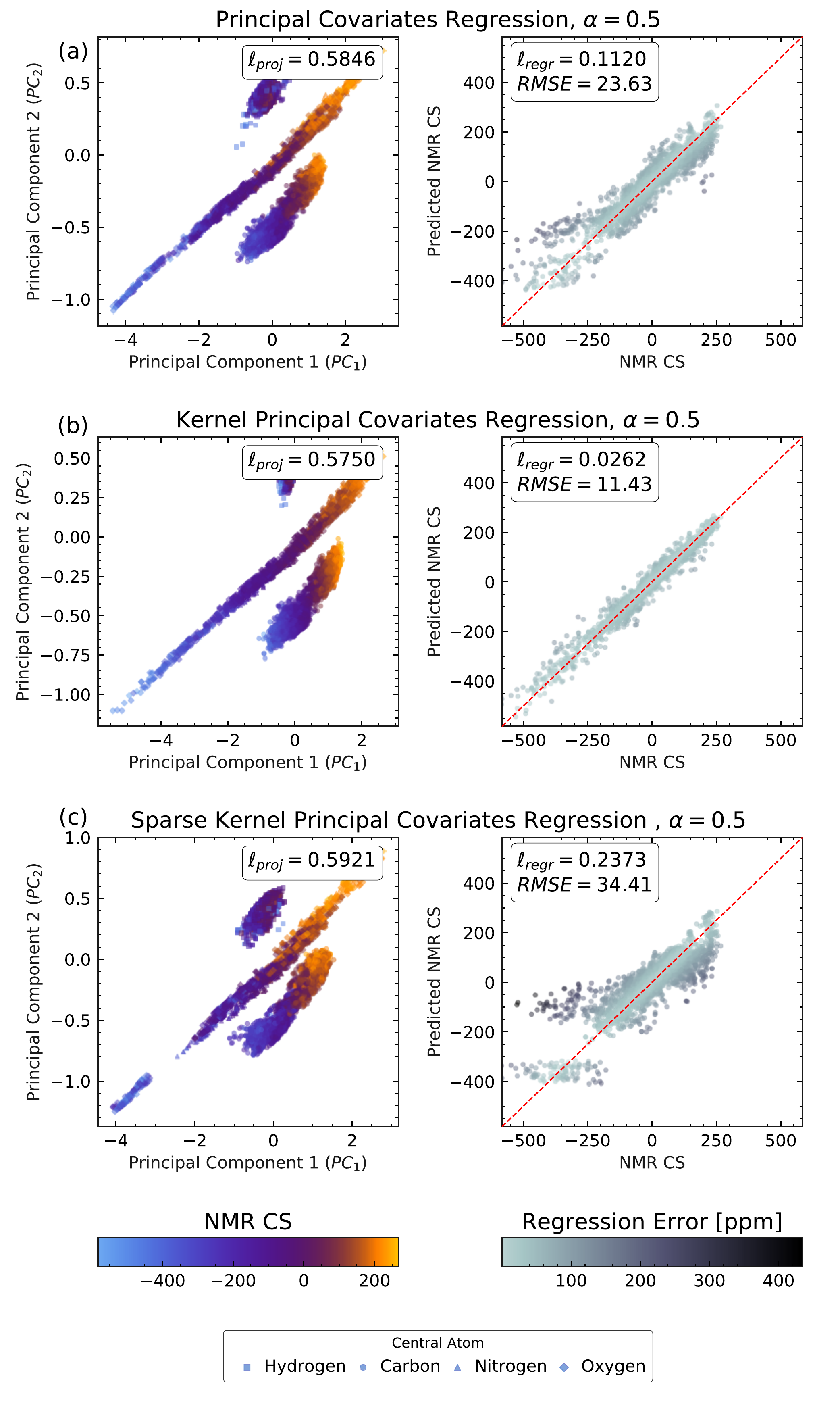}
    \caption{\textbf{Projections and Regression from PCovR Models.} Analogous to Fig.~\ref{fig:models}, we show the loss incurred from the projections and regression of the three PCovR models at $\alpha=0.5$. (a) Principal Covariates Regression (PCovR), (b) Kernel PCovR (KPCovR), and (c) Sparse KPCovR with $n_{active}=50$. $\alpha$ was chosen to best compare the models, although ideal $\alpha$ may fluctuate between models, albeit often with a range of suitable $\alpha$, as in Fig.~\ref{fig:PCovR}. The inset includes the  $\ell_\text{proj} = \lproj$ and $\ell_\text{regr} = \lregr$.}
    \label{fig:kpcovr-models}
\end{figure}

\subsubsection{Sparse Kernel Principal Component Analysis}
\label{sec:skpca}

We can define the covariance in the kernel feature space along with its eigendecomposition,
\begin{equation}
    \bC = \bPhi_{\N\M}^T\bPhi_{\N\M} = \bU_\bC\bLAM_\bC\bU_\bC^T,
\end{equation}
and subsequently compute the projections analogously to standard KPCA
\begin{equation}
    \bT = \bPhi_{\N\M}\hat{\bU}_\bC = \bG_{\N\M}\bU_{\bG_{\M\M}}\bLAM_{\bG_{\M\M}}^{-1/2}\hat{\bU}_\bC=\bG_{\N\M}\bPkt,
\end{equation}
which effectively determine the directions of maximum variance of the samples in the active RHKS.

Fig.~\ref{fig:models}e shows that with an active set size of just 50 samples (out of more than 12,000), selected by FPS~\cite{imba+18jcp}, one can obtain a KPCA latent projection that  matches very well the qualitative features of the full KPCA construction.

\subsubsection{Sparse Kernel Ridge Regression}
\label{sec:skrr}

In sparse KRR we proceed as in standard KRR, but use the feature matrix from the Nystr\"om approximation. The corresponding regularised LR loss in the kernel feature space is
\begin{equation}
    \ell = \loss{\bY}{\bPhi_{\N\M}\bPpy} + \lambda\lVert\bPpy\rVert^2
\end{equation}
for which the solution is
\begin{equation}
    \begin{split}
    \bPpy &= \left(\bPhi_{\N\M}^T\bPhi_{\N\M} + \lambda\bI\right)^{-1}\bPhi_{\N\M}^T\bY \\
    &= \left(\bPhi_{\N\M}^T\bPhi_{\N\M} + \lambda\bI\right)^{-1}\bLAM_{\bG_{\M\M}}^{-1/2}\bU_{\bG_{\M\M}}^T\bG_{\N\M}^T\bY.
    \end{split}
\end{equation}
Alternatively, we can redefine the weights so that
\begin{equation}
   \bYhat = \bPhi_{\N\M}\bPpy = \bG_{\N\M}\bU_{\bG_{\M\M}}\bLAM_{\bG_{\M\M}}^{-1/2}\bPpy = \bG_{\N\M}\bPky,
\end{equation}
from which we see that
\begin{equation}
    \begin{split}
        \bPky &= \bU_{\bG_{\M\M}}\bLAM_{\bG_{\M\M}}^{-1/2}\bPpy \\
              &= \bU_{\bG_{\M\M}}\bLAM_{\bG_{\M\M}}^{-1/2}\left(\bPhi_{\N\M}^T\bPhi_{\N\M} + \lambda\bI\right)^{-1} \\
              &\times\bLAM_{\bG_{\M\M}}^{-1/2} \bU_{\bG_{\M\M}}^T\bG_{\N\M}^T\bY.
    \end{split}
\end{equation}
By writing out explicitly $\bPhi_{\N\M}^T\bPhi_{\N\M}$ in terms of $\bG_{\N\M}$ we obtain\cite{smola_sparse_2000}
\begin{equation}
    \bPky = \left(\bG_{\N\M}^T\bG_{\N\M} + \lambda\bG_{\M\M}\right)^{-1}\bG_{\N\M}^T\bY.
\end{equation}
As shown in Fig.~\ref{fig:models}f, an active set size of 50 is not sufficient to achieve an accurate regression model, and the error is larger than with a linear regression method. However, the error can be reduced systematically by increasing the size of the active set, finding the best balance between accuracy and cost (see SI).

\section{Extensions to Principal Covariates Regression}
\label{sec:extensions}

After having summarised existing linear and kernel methods for feature approximation and property prediction, we now introduce kernelized PCovR (KPCovR), as a way to combine the conceptual framework of PCovR and the non-linear features afforded by a kernel method.

\subsection{Full kernel PCovR}
\label{sec:kpcovr}

We start by constructing the augmented kernel matrix as a combination of KPCA and KRR. In particular, we substitute $\bPhi$ for $\bX$ and the KRR solution of $\bY$, $\bYhat = \bG\left(\bG + \lambda\bI\right)^{-1}\bY$, for $\bY$, so that we have
\begin{equation}
    \bGt = \alpha\bG + (1 - \alpha)\bYhat\bYhat^T, \label{eq:pcovr-kernel}
\end{equation}
where we consider the kernel matrix to be standardised in a way that is equivalent to normalising $\bPhi$ (see Appendix~\ref{app:centring}).
Just as in PCovR, the unit variance projections $\bTt$ are given by the top eigenvectors $\hat{\bU}_{\bGt}$ of $\bGt$, and the non-whitened projections as $\bT = \hat{\bU}_{\bGt} \hat{\bLAM}_{\bGt}^{1/2}= \bGt\hat{\bU}_{\bGt} \hat{\bLAM}_{\bGt}^{-1/2} $, corresponding to the RKHS $\tilde{\bPhi}$ associated with the PCovR kernel [Eq.~\eqref{eq:pcovr-kernel}].

Projecting a new set of structures in the kernel PCovR space entails computing the RHKS between the samples that were originally used to determine the KPCovR features and the new samples. Given that one may not want to compute these explicitly, it is useful to define a projection acting directly on the kernel, such that $\bT=\bG \bPgt$:

\begin{equation}
    \begin{split}
    &\bPgt = \\
    &\left(\alpha\bI + (1 - \alpha)\left(\bG + \lambda\bI\right)^{-1}\bY\bYhat^T\right) \hat{\bU}_{\bGt} \hat{\bLAM}_{\bGt}^{-1/2}.
    \end{split}
\end{equation}
We also determine the matrix that enables predictions of properties from the latent space $\bT$ through LR, just as in the linear case [Eq.~\eqref{eq:Pty}].
Computing the projection loss minimised by KPCovR, $\ell_\text{proj} = \lprojphi$, is trivial if one computes explicitly a RKHS approximation of $\bPhi$, but it requires some work if one wants to avoid evaluating $\bPhi$ (see Appendix~\ref{app:kpcovr-loss}).

As shown in Fig.~\ref{fig:KPCovR}, the method combines a behaviour similar to linear PCovR with the improved property prediction accuracy afforded by kernel methods.
In the the low-$\alpha$ regime the regression accuracy approaches that of KRR, and the projection accuracy converges to a KPCA-like behaviour for $\alpha\approx 1$. For the optimal value of $\alpha$, the latent-space map (shown in Fig.~\ref{fig:kpcovr-models}b) combines a clear separation of structurally-distinct clusters with a 70\% reduction in regression error when compared to linear PCovR ($\ell_\text{regr} = 0.026$ vs. $\ell_\text{regr} = 0.112$).

\subsection{Sparse Kernel PCovR}
\label{sec:skpcovr}

Our derivation of the sparse version of KPCovR can be obtained almost directly from that of feature-space PCovR by taking explicitly the projection of the kernel on the active RKHS $\bPhi_{\N\M}=\bG_{\N\M} \bU_{\bG_{\M\M}} \bLAM_{\bG_{\M\M}}^{-1/2}$.
One can then define the covariance of the active kernel features
\begin{align}
    \bC &=  \bPhi_{\N\M}^T\bPhi_{\N\M} \\&=
    \bLAM_{\bG_{\M\M}}^{-1/2} \bU_{\bG_{\M\M}}^T \bG_{\N\M}^T
    \bG_{\N\M} \bU_{\bG_{\M\M}} \bLAM_{\bG_{\M\M}}^{-1/2},
\end{align}
and use it in the definition of the modified KPCovR covariance

\begin{equation}
    \begin{split}
    \bCt =& \alpha\bC \\+& (1 - \alpha)\bC^{1/2}\left(\bC + \lambda\bI\right)^{-1}
    \bLAM_{\bG_{\M\M}}^{-1/2} \bU_{\bG_{\M\M}}^T \bG_{\N\M}^T \bY \\
    \times &\bY^T \bG_{\N\M} \bU_{\bG_{\M\M}} \bLAM_{\bG_{\M\M}}^{-1/2}
    \left(\bC + \lambda\bI\right)^{-1}\bC^{1/2}.
    \end{split}
\end{equation}

With these definitions, the projection matrices onto the (sparse) KPCovR latent space and onto the latent-space-restricted properties, analogous to Eq.~\eqref{eq:pcovr-projectors}, read
\begin{equation}
    \begin{split}
    \bPgt =&\bU_{\bG_{\M\M}} \bLAM_{\bG_{\M\M}}^{-1/2}\bC^{-1/2}\hat{\bU}_{\bCt}\hat{\bLAM}_{\bCt}^{1/2} \\
    \bPty =& \hat{\bLAM}_{\bCt}^{-1/2} \hat{\bU}_{\bCt}^T  \bC^{-1/2} \bLAM_{\bG_{\M\M}}^{-1/2} \bU_{\bG_{\M\M}}^T \bG_{\N\M}^T\bY. \\
    \end{split}
    \label{eq:skpcovr-projectors}
\end{equation}
Similar to what we observed for sparse KPCA and KRR, reducing the active space to 50 active samples preserves the qualitative features of the latent-space map, but leads to substantial loss of performance for regression (Fig.~\ref{fig:kpcovr-models}c). The error, however, is equal to that observed for sparse KRR, which indicates that it is due to the limited  active space size, and not by the dimensionality reduction.

\begin{table*}[tbhp]
\centering
\begin{tabular}{c|ccc|cccc|cccc}
\multicolumn{4}{c}{} & \multicolumn{4}{c}{{\textbf{PCovR}}} & \multicolumn{4}{c}{{\textbf{KPCovR}}} \\ 
 \hline\hline
\textbf{Dataset} & $\npca$ & $N$ & $\ell$ & $\alpha = 0.0$ & $\alpha = \alpha^*$ & $\alpha = 1.0$ &$\alpha^*$ &  $\alpha = 0.0$ & $\alpha = \alpha^*$ & $\alpha = 1.0$ & $\alpha^*$\\ 
 \hline
\multirow{2}{*}{\textbf{CSD-1000R}} & \multirow{2}{*}{2} & \multirow{2}{*}{25600} & $\ell_\text{regr}$ & 0.1106 & 0.112 & 0.9285 & \multirow{2}{*}{0.5} & 0.0249 & \bf 0.0262 & 0.9285 & \multirow{2}{*}{0.5}\\ 
 & & & $\ell_\text{proj}$ & 0.9633 & 0.5846 & 0.4586 & & 0.9656 & 0.575 & 0.4439 & \\ 
 \hline
\multirow{2}{*}{\textbf{C-VII}} & \multirow{2}{*}{2} & \multirow{2}{*}{10874} & $\ell_\text{regr}$ & 0.0707 & 0.0753 & 0.9447 &\multirow{2}{*}{0.5} & 0.0133 & \bf 0.0172 & 0.9445 &\multirow{2}{*}{0.5} \\ 
 & & & $\ell_\text{proj}$ & 0.9572 & 0.4443 & 0.2384 & & 0.9607 & 0.4515 & 0.2398 & \\ 
 \hline
\multirow{2}{*}{\textbf{Deem} (global)} & \multirow{2}{*}{2} & \multirow{2}{*}{4000} & $\ell_\text{regr}$ & 0.0686 & 0.121 & 0.6084 &\multirow{2}{*}{0.5} & 0.0586 & 0.1315 & 0.6071 &\multirow{2}{*}{0.55} \\ 
 & & & $\ell_\text{proj}$ & 0.5783 & 0.4287 & 0.2441 & & 0.5683 & 0.3897 & 0.2149 & \\ 
 \hline
\multirow{2}{*}{\textbf{Deem} (local)} & \multirow{2}{*}{2} & \multirow{2}{*}{10968} & $\ell_\text{regr}$ & 0.365 & 0.0841 & 0.7279 &\multirow{2}{*}{0.5} & 0.0015 & \bf 0.0464 & 0.7163 &\multirow{2}{*}{0.5} \\ 
 & & & $\ell_\text{proj}$ & 0.7623 & 0.6531 & 0.4258 & & 0.7652 & 0.6648 & 0.4577 & \\ 
 \hline
\multirow{2}{*}{\textbf{QM9}} & \multirow{2}{*}{2} & \multirow{2}{*}{10000} & $\ell_\text{regr}$ & 0.3298 & 0.3952 & 0.4789 &\multirow{2}{*}{0.45} & 0.3135 & 0.3891 & 0.4789 &\multirow{2}{*}{0.5} \\ 
 & & & $\ell_\text{proj}$ & 0.7419 & 0.5588 & 0.5361 & & 0.6281 & \bf 0.3677 & 0.3407 & \\ 
 \hline
\multirow{2}{*}{\textbf{QM9}} & \multirow{2}{*}{12} & \multirow{2}{*}{10000} & $\ell_\text{regr}$ & 0.1212 & 0.1296 & 0.2938 &\multirow{2}{*}{0.4} & 0.0744 & \bf 0.083 & 0.2938 &\multirow{2}{*}{0.55} \\ 
 & & & $\ell_\text{proj}$ & 0.4511 & 0.3493 & 0.3287 & & 0.2686 & \bf 0.0822 & 0.0459 & \\
 \hline
 \multirow{2}{*}{\textbf{Arginine Dipeptide}} & \multirow{2}{*}{2} & \multirow{2}{*}{4217} & $\ell_\text{regr}$ & 0.0121 & 0.013 & 0.6067 &\multirow{2}{*}{0.55} & 0.0031 & \bf 0.0038 & 0.6006 &\multirow{2}{*}{0.5} \\ 
 & & & $\ell_\text{proj}$ & 0.8237 & 0.549 & 0.435 & & 0.827 & 0.5741 & 0.4655 & \\ 
 \hline
\multirow{2}{*}{\textbf{Azaphenacenes}} & \multirow{2}{*}{2} & \multirow{2}{*}{311} & $\ell_\text{regr}$ & 0.4632 & 0.5537 & 0.8929 &\multirow{2}{*}{0.6} & 0.5181 & 0.5582 & 0.8834 &\multirow{2}{*}{0.65} \\ 
 & & & $\ell_\text{proj}$ & 0.8295 & 0.5113 & 0.3742 & & 0.8583 & 0.4689 & 0.3124 & \\ 
 \hline\hline
\end{tabular}
\caption{\textbf{Performance of PCovR and KPCovR} for the different examples using $\ell_\text{regr} = \lregr$ and $\ell_\text{proj} = \lproj$. The $\alpha$ which minimizes the total loss $\ell = \ell_\text{regr} + \ell_\text{proj}$ is given by $\alpha^*$. Results of optimal-alpha KPCovR are always comparable or better than those from the linear PCovR version. Values that are improved by more than 10\% are highlighted in bold face.}
\label{table:perf}
\end{table*}

\section{Examples}
Up until this point, we have talked about the definition of KPCovR in an abstract, equations-heavy manner. Here, we will demonstrate the usage of such method for a wide range of materials science datasets. These datasets have all been already published elsewhere, and we leave to the SI a precise discussion of their structure, content and provenance, as well as a thorough analysis of the behaviour of the different linear and kernel methods applied to each data set. Here we limit ourselves to the most salient observations, and summarise the insights that could be relevant to the application of (K)PCovR to other materials and molecular datasets. We also distribute data files in the supplementary information that can be viewed with the interactive structure-property explorer chemiscope\cite{chemiscope}. We encourage the reader to use them to gain a more interactive support to follow the discussion in this section.

For each dataset, we trained machine learning models on a randomly-chosen half of the included samples, and then evaluated these models on the remaining samples. In the following section, we report losses, errors and figures on the validation set of points only. The total number of samples we considered are available in Table \ref{table:perf}, together with performance metrics that show that PCovR-like methods achieve consistently an excellent compromise between the optimisation of $\ell_\text{proj}$ and $\ell_\text{regr}$, and demonstrate that KPCovR outperforms by a large margin its non-kernelized counterpart in terms of regression performance.

\begin{figure}[tbhp]
   \centering
   \includegraphics[width=\linewidth]{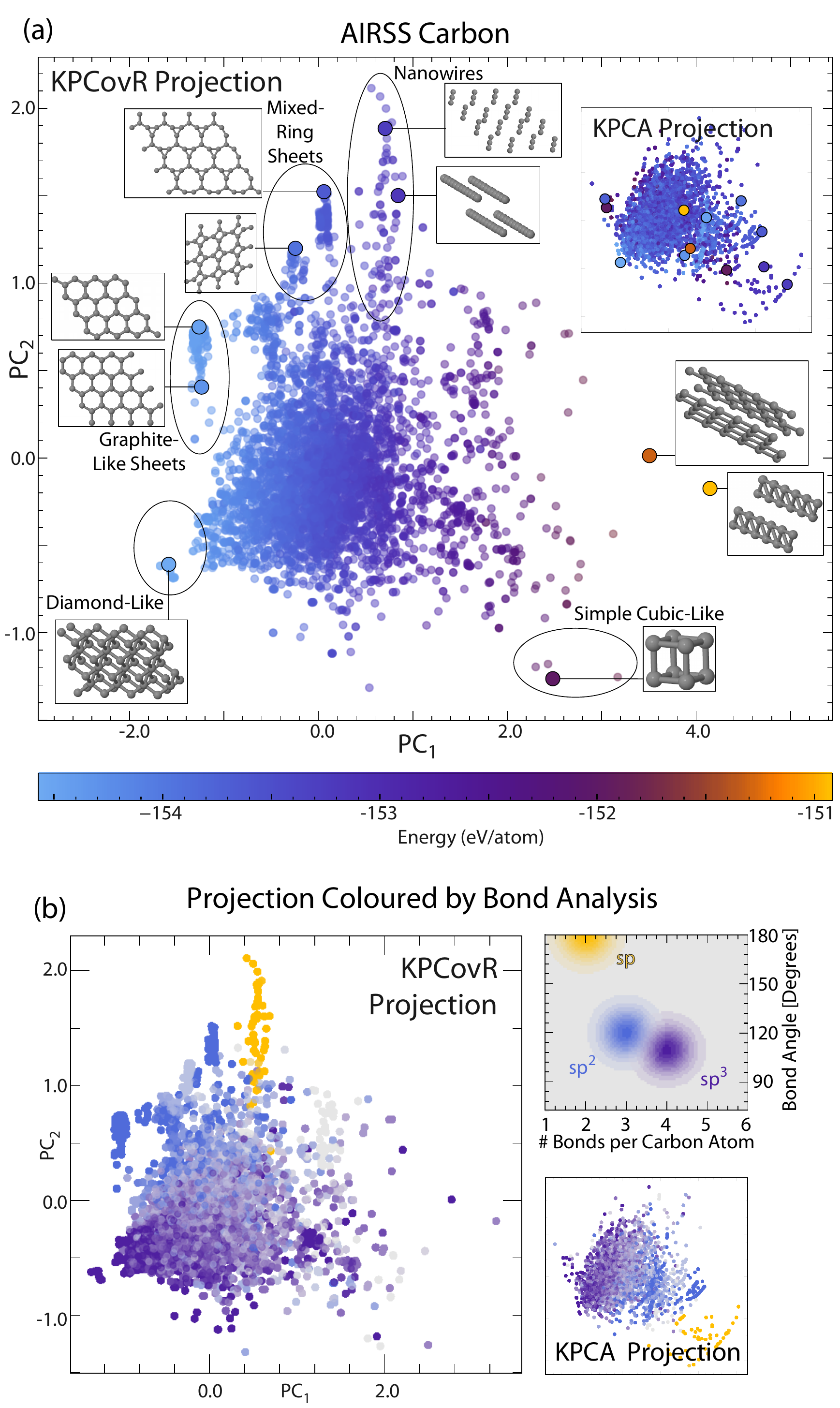}
   \caption{\textbf{KPCovR for AIRSS Carbon Structures and Energies.} (a) Projections coloured by the per-atom energy. KPCovR projection at $\alpha = 0.5$. Clusters identified using $chemiscope$ have been labelled with representative snapshots provided. The KPCA projection is given in the upper right inset, with highlighted structures in the KPCovR projection denoted by enlarged points. (b) Projections in (a) recoloured by the bond analysis, where yellow, blue and purple denote similarity to $sp$, $sp^2$, and $sp^3$ geometries, respectively, and grey signifying no resemblance.}
   \label{fig:carbon}
\end{figure}

\subsection{Carbon}
We apply KPCovR to the C-VII carbon dataset, which contains roughly 11,000 carbon structures generated using Ab Initio Random Structure Searching (AIRSS) at 10GPa \cite{pick-need11jpcm, pickard-carbon}. Here, a KRR model predicts the average per-atom energy of each structure with a RMSE of 0.055 eV/atom
(equivalent to $\ell_\text{regr}=0.0707$) yet can only describe 4\% of the latent space variance (\textit{i.e.} $\ell_\text{proj} = 0.96$). The KPCA model retains 76\% of the latent space variance but with a RMSE of 0.46 eV/atom ($\ell_\text{proj} = 0.9447$). By comparison, at the optimal $\alpha$ value, KPCovR sacrifices little in regression accuracy (0.062 eV/atom, $\ell_\text{regr} = 0.0753$) and latent space variance retention (55\%). 

Additionally, KPCovR provides a more intuitive qualitative picture for understanding the dataset.
In the original KPCA projection, the principal components correlate strongly with the dimensionality of the carbon structures, with linear nanowires in the lower right of the projection, sheets and planar structures above to the left, and 3D structures conglomerated in the upper left. The KPCovR projection does not only show a much clearer correlation between the position on the map and the stability of each configuration, but also provides more compact clustering of similar structures  (Fig.~\ref{fig:carbon}(a)).
The nanowires  (typically linear carbon bonds) are located in the upper centre, the planar structures are partitioned in smaller clusters to the upper left, with clear sub-clusters associated with different ring patterns, and 3D crystals are distributed throughout the lower left and centre, with a well-separated cluster for tetrahedral carbon.
The structural homogeneity of the clusters is confirmed by bond analysis -- typically, the most common environments found in low energy 1D, 2D, and 3D carbon structures correspond to $sp$, $sp^2$, and $sp^3$ geometry. Angles between neighbouring bonds typically serve as a good proxy for detecting this environments, where the ideal values are $180^\circ$, $120^\circ$, $109.5^\circ$, respectively. Both KPCA and KPCovR detect clusters delineated by the number of bonds and bond angles (Fig.~\ref{fig:carbon}(b)), with these clusters arranged right-to-left in the KPCA projection, and top-to-bottom in the KPCovR projection.
However, in the KPCovR projection, there is another gradient visible, with structures to the left of the projection more strongly coinciding with the ideal $sp$, $sp^2$, and $sp^3$, and those to the right being increasingly distorted.\footnote{The two notable exceptions--a dark purple cluster middle centre and a grey cluster upper left-- can be shown to be associated with bond angle distributions which are multimodal, leading to incorrect classification by a criterion based on the \textit{mean} bond angles.}
Thus, bond analysis reveals that, in addition to the clear delineation between structure dimensionality provided by KPCA, the inclusion of an energy regression criterion in the KPCovR loss leads to a map that more closely coincides with the conventional understanding of stable structures as those that have low distortion relative to the ideal carbon bonding geometries.

\subsection{Zeolites}

\begin{figure*}
    \centering
    \includegraphics[width=\linewidth]{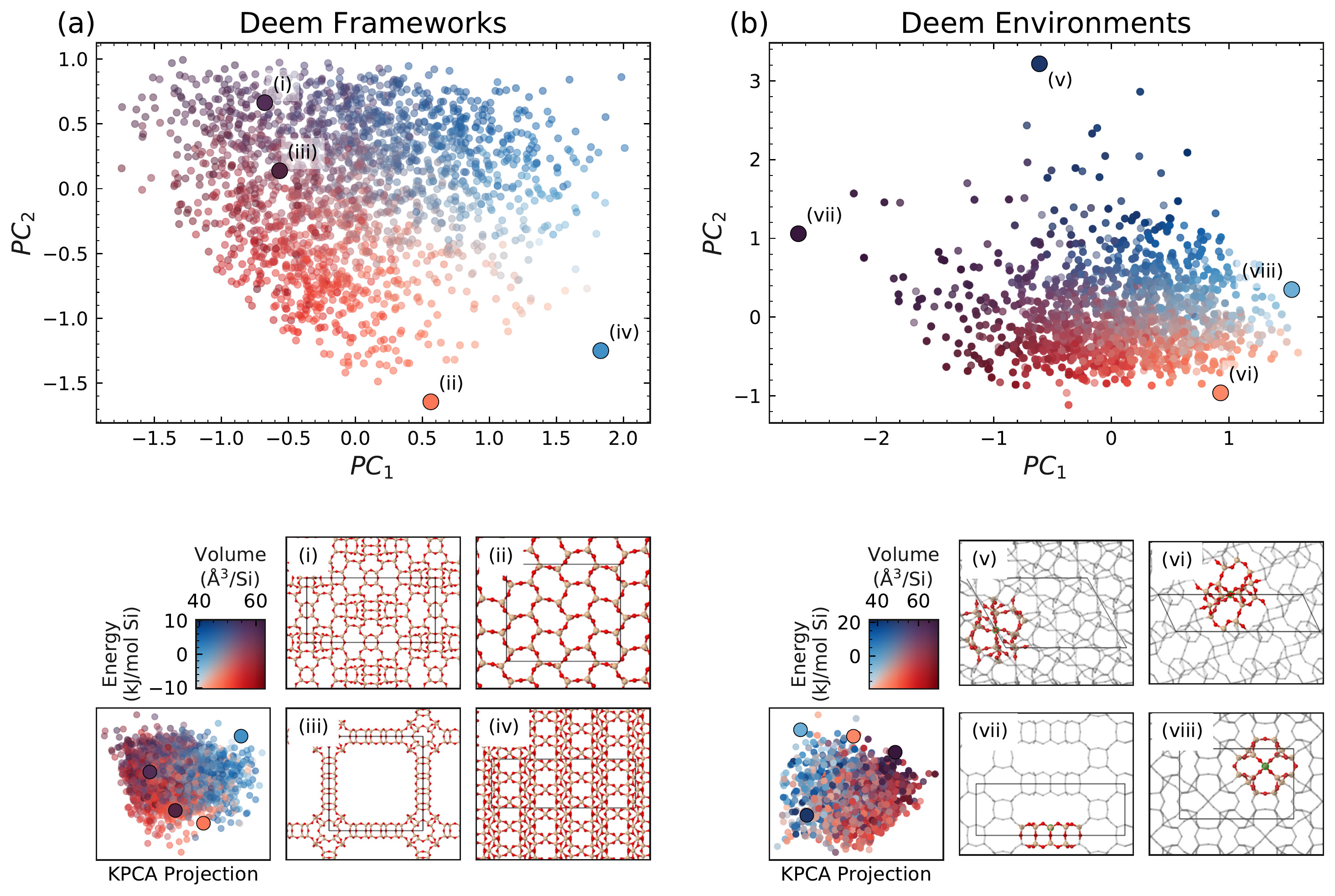}
    \caption{\textbf{KPCovR for Zeolite Structures and Environments.} Two-dimensional KPCovR projections at the optimal $\alpha$ for (a) 2000 zeolite structures and (b) 5484 zeolite environments from the Deem SLC PCOD database. The points in the projections are coloured by the molar energy and volume according to the 2D colour map below the projection. For clarity, energies are centred relative to the the mean energy per \ce{Si} over all 10000 structures from Ref. \citenum{helf+19jcp}. The frameworks and environments with the highest and lowest volumes and energies are also pictured below their respective projections. Below each KPCovR projection is an analogous KPCA to facilitate a qualitative comparison with the KPCovR projections.
    }
    \label{fig:deem_summary}
\end{figure*}
%
%
We apply KPCovR to a subset of the Deem data set of hypothetical silica zeolites~\cite{Pophale2011}, where the use of KPCA to construct an atlas of the building blocks of a zeolite was previously demonstrated by the authors~\cite{helf+19jcp}.
By construction all frameworks in the dataset are based on tetrahedrally coordinated \ce{SiO4} units yet differ considerably in terms of molar energies and volumes.
A KRR model based on an additive combination of environment Gaussian kernel, built using atom-centred  SOAP features with a cutoff of 6.0\AA{} achieves an excellent accuracy in predicting the lattice energy (with an error around 1.88 kcal/mol), and molar volume (with an error around 1.97 cm$^3$/mol), with $\ell_\text{regr} =0.0686$.
However, the first 2 KPCA components correlate rather weakly with these properties. A data representation based on those provides information on the structural diversity, but describes only qualitatively structure-property relations.
As shown in Table~\ref{table:perf} KPCovR at the optimal $\alpha$ provides a much more effective description; the latent space covers 61\% of the structural diversity, while providing enough information to predict accurately lattice energy (2.64 kcal/mol) and molar volume (2.37 cm$^3$/mol), $\ell_\text{regr}=0.13$.
The map naturally orders structures between regular frameworks that have intermediate densities and low lattice energy \ref{fig:deem_summary}(a,iii)., to the bottom, to frameworks with very large pores that have very low density and usually intermediate of high lattice energies. While there are no clear clusters emerging (which is not unexpected given the origin of the dataset as a high-throughput, random search) one can often observe that nearby structures exhibit similar structural motifs. For instance, most of the structures on the top left side of the map are associated with large 1D channels, as shown in Fig. \ref{fig:deem_summary}(a,iii).

For a system exhibiting a combinatorial number of metastable structures, such as silica frameworks, an analysis based on the structural building blocks is often more insightful than the analysis of the overall structures. When using atom-centred features, or additive kernels built on them, it is natural to regard additive properties such as volume or energy as arising from a sum of environmental contributions, and to use these atom-centred environments as the building blocks to rationalise structure-property relations.
As discussed in appendix~\ref{app:local-global}, the construction of a regression model for the framework properties yields as a side-effect a data-driven partitioning, that can be used for a (K)PCovR analysis of such building blocks. 
The resulting representation  (Fig.~\ref{fig:deem_summary}(b)) shows an excellent correlation between the position in a 2D representation and the predicted contributions to lattice energy and molar volume (site energy RMSE: 3.18 kJ/mol Si, site volume RMSE: 3.67 kJ/mol Si, 34\% of structural variance). 
As observed in Ref.~\citenum{helf+19jcp}, and consistent with what is seen in the framework analysis, the thorough search of potential frameworks that underlies the construction of this dataset is reflected in the lack of substantial clustering of the environments. 
The bulk of the latent space is uniformly covered, and one does not see an obvious qualitative relation between properties and the local topology of the framework. Regular structures are mapped side-by-side to disordered environments.
Only at the extremes one can recognise clearer patterns. (A few extremal structures and their location in the KPCovR projection are highlighted in Fig. \ref{fig:deem_summary}(b).) High-energy (and hence poorly stable) building blocks can be distinguished between high-density structures that contain highly-strained 3-fold rings, and low-density structures, associated to the surface of a large cavity, and to `pillar-like'' motifs that are present in the most highly porous frameworks (Fig.~\ref{fig:deem_summary}(b,vii)).
Low-energy structures, in the lower side of the map, tend to have low and intermediate volume, and are predominantly associated with six- and four-member rings.

These are however weak correlations, and in general there are no apparent patterns that correlate the framework topology and the position on the map. This confirms, in a more direct manner, the structure-property insights that were inferred by \emph{separate} application of supervised and unsupervised algorithms in Ref.~\citenum{helf+19jcp} -- namely that, for four-coordinated silica frameworks, the topology of the network is a good predictor of energy and density only for extremes. For the bulk of the possible binding motifs, a low-dimensional representation is not sufficient to capture the extreme structural diversity, and to rationalise the multitude of alternative building blocks that give rise to similar macroscopic materials properties.

\subsection{QM9}

\begin{figure*}[tbhp]
    \centering
    \includegraphics[width=0.8\linewidth]{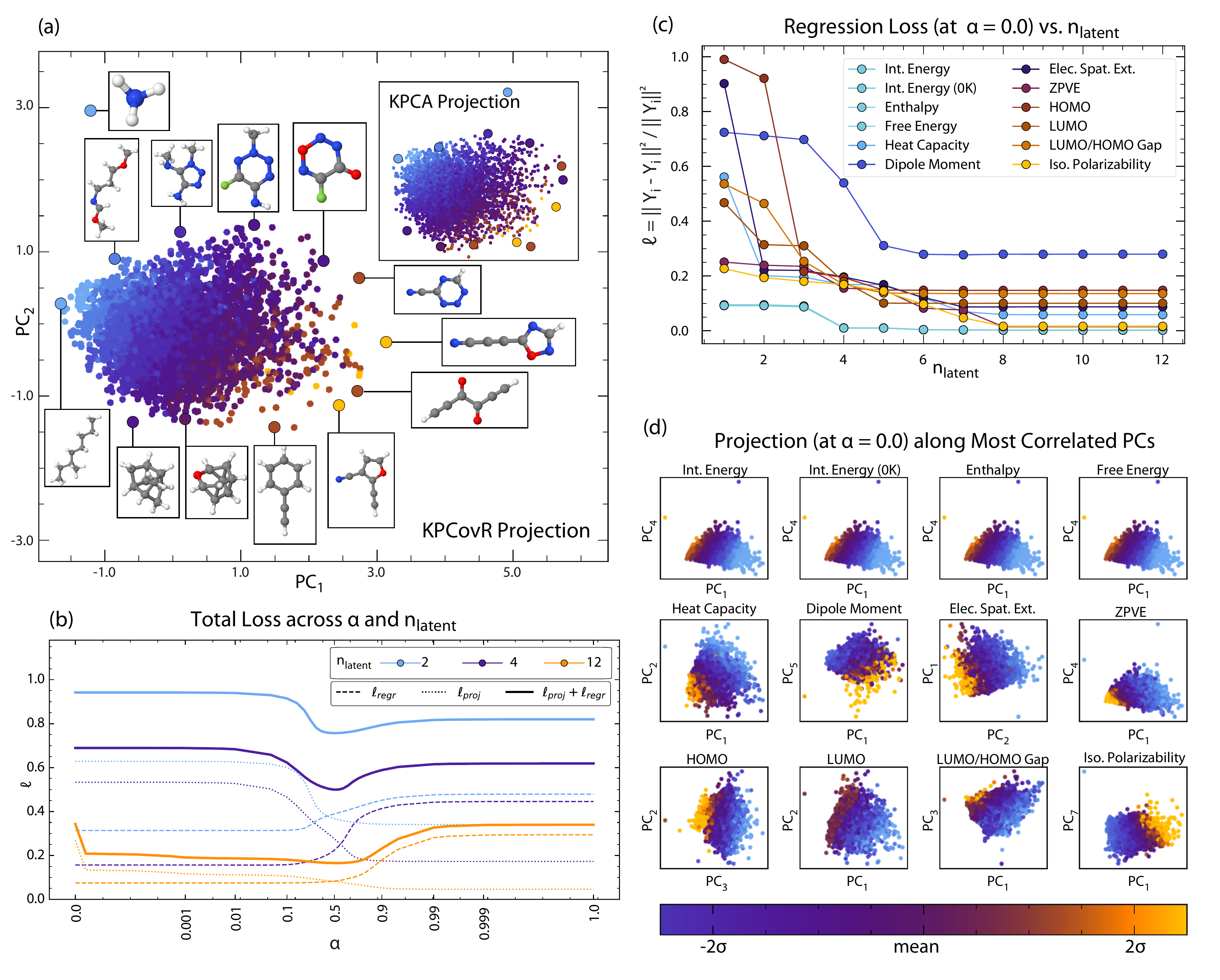}
    \caption{\textbf{KPCovR for the QM9 Dataset.} (a) Comparison of KPCovR ($\alpha = 0.5$) and KPCA (inset) projections, plotted along the first two principal components. Several molecules have been highlighted in insets. (b) Total loss across $\alpha$ and the number of principal components for 2 (blue), 4 (purple), and 12 (orange) principal components. Solid lines denote total loss $\ell$, dashed $\ell_\text{regr}$ and dotted $\ell_\text{proj}$. (c) Per-property regression loss at $\alpha = 0$ for the 12 properties included. Note that the internal energies, enthalpy, and free energy fully overlap and are given by a single colour. (d) Projections at $\alpha=0.0$ of the 12 properties along their most correlated principal components, where each projection is coloured by the corresponding property, scaled in accordance with the colour bar below.
    }
    \label{fig:qm9_summary}
\end{figure*}

Our next case study regards the QM9 dataset, which contains over 133,000 molecules consisting of carbon, hydrogen, nitrogen, oxygen, and fluorine \cite{qm9-ramakrishnan, qm9-data}, of which we use 10,000 for this study. To demonstrate the application of KPCovR to multi-target learning, we construct our models using all the 12 properties available in the dataset: internal energy at 0K and 298.15K, free energy at 298.15K, enthalpy at 298.15K, highest occupied molecular orbital (HOMO), lowest unoccupied molecular orbital (LUMO), HOMO-LUMO gap, heat capacity, zero point vibrational energy (ZPVE), dipole moment, isotropic polarisability, electronic spatial extent (ESE).

Here, there is not a large qualitative difference between the two-dimensional KPCovR projection at intermediate $\alpha$ and that constructed via KPCA, with the former retaining 63\% of the variance compared to the latter's 66\%. The energetic properties, which are well-represented by these first two principal components, are also well-correlated with the degree of unsaturation, and thus the structural diversity of the dataset. This is summarised in Fig. \ref{fig:qm9_summary}(a), where the projections for $\alpha = 0.5$ and $\alpha = 1.0$ are shown, coloured by the degree of unsaturation\footnote{The degree of unsaturation, defined as $d = C-{\frac {H}{2}}-{\frac {X}{2}}+{\frac {N}{2}}+1$, where X is a halogen, estimates the number of rings and $\pi$ bonds in the molecule.} and with representative molecules highlighted. Besides the left-to-right saturated-to-unsaturated trend, the map position also correlates with the presence of O, N, F atoms, that increase from bottom to top.

\begin{figure*}
    \centering
    \includegraphics[width=\linewidth]{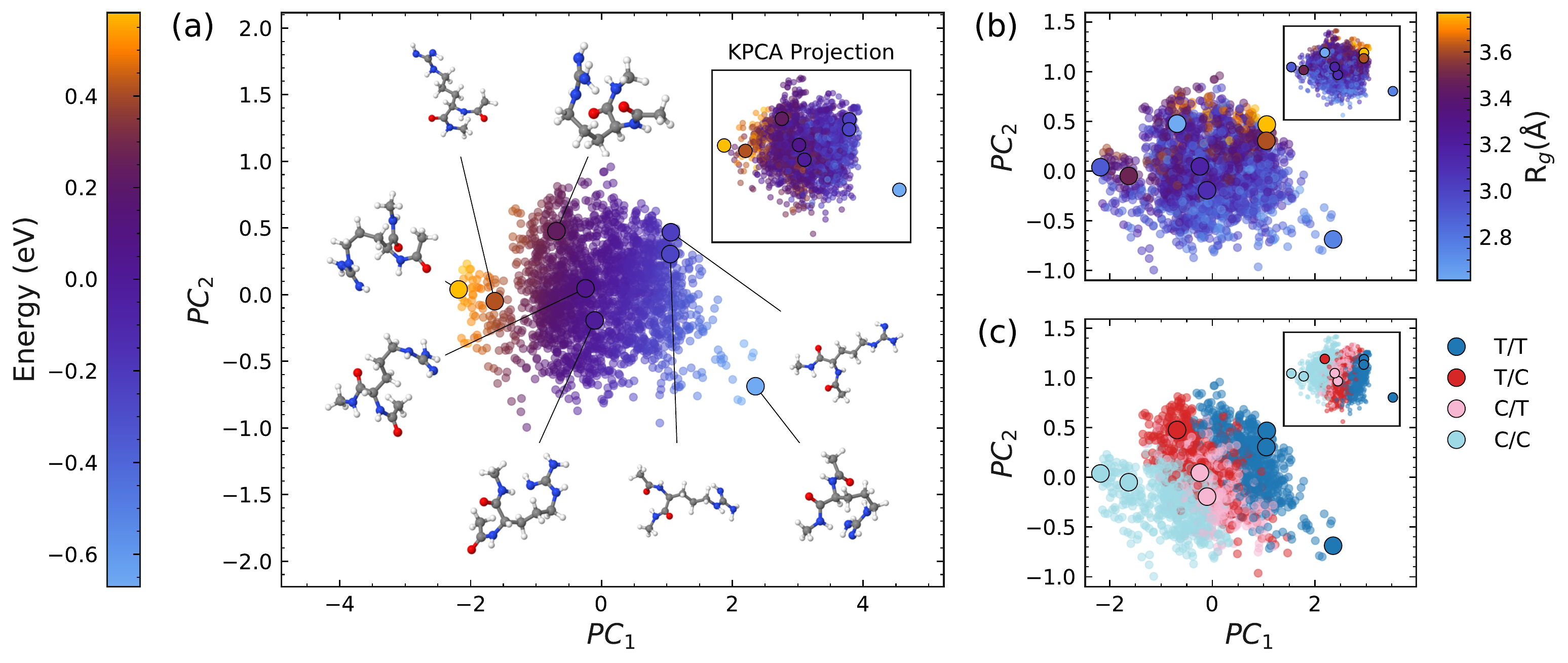}
    \caption{\textbf{KPCovR of Arginine Dipeptides.} Two-dimensional KPCovR projections at the optimal $\alpha$ for bare arginine dipeptides coloured by the (a) structure energies (centred relative to mean energy over all 4217 structures), (b) radii of gyration ($R_g$), and (c) peptide bond isomerism. The peptide bond isomerism is determined by the dihedral angle $\omega$ of the two peptide bonds. A particular peptide bond is classified as cis (C) if $-\pi/2 < \omega < \pi/2$, and the bond is classified as trans (T) if $-\pi < \omega < -\pi/2$ or $\pi/2 < \omega < \pi$. For each KPCovR projection, the inset contains the corresponding KPCA projection. In each KPCovR and KPCA projection, several points are highlighted, and their corresponding atomic snapshots are shown in (a).
    }
    \label{fig:arginine_summary}
\end{figure*}

Due to the large number of properties used as targets, a low-dimensional latent space cannot achieve the same prediction accuracy as (kernel) ridge regression, for $\alpha=0$ and by extension at intermediate values of $\alpha$. It is necessary to retain a larger number of latent space components to obtain a model capable of effective regression, as seen in Table \ref{table:perf}, where $\ell_\text{regr}$ goes from 0.31 with $\npca=2$ to 0.07 with $\npca=12$, and seen in Fig.~\ref{fig:qm9_summary}(b). 
For a given value of $\alpha$, both regression and projection errors are bound to decrease when retaining a larger number of PCs. The optimal value of $\alpha$, however, is not necessarily the same for increasing numbers of PCs particularly in datasets where $\ell_\text{regr}$ and $\ell_\text{proj}$ have a magnitude that varies with $\npca$ in a different way. In this case, however, the optimal $\alpha$ is nearly constant, as shown in Fig.~\ref{fig:qm9_summary}(b). The figure also shows a sudden drop in $\ell_\text{proj}$ for $\alpha > 0$. This  discontinuity in the variance suggests that insufficient information is contained in the 12 properties to construct an orthogonal set of 12 principal components, and thus some properties must be highly correlated.

Models constructed with fewer principal components can provide insight into the nature of the properties included, particularly in the cases weighted towards regression as $\alpha \to 0$. 
In Fig.~\ref{fig:qm9_summary}(c), we show the regression errors of the individual properties as a function of $\npca$. 
For each property, the decay of $\ell_\text{regr}$ when incorporating a new principal component indicates how strongly the new feature and the property are correlated, and gives indirect information on the correlation between properties.
For instance, we can see that (unsurprisingly) the internal energies, enthalpy and free energy are heavily correlated with each other, as their associated $\ell_\text{regr}$ decreases precisely in the same way, indicative of a strong correlation the first and fourth principal components.
Fig.~\ref{fig:qm9_summary}(d) shows color-coded maps of the 12 targets, using for each of them the two PCs that lead to the largest decrease in $\ell_\text{regr}$.
%

\subsection{Arginine Dipeptide}

We also applied KPCovR to the 4219 arginine dipeptide conformers that are collected in the Berlin amino acid database \cite{ropo_first-principles_2016}, that was also investigated using a purely unsupervised dimensionality-reduction scheme~\cite{de+16jci}. The conformer energy was used as the target property for the purpose of constructing the KPCovR model. Fig. \ref{fig:arginine_summary} shows the two-component KPCovR projection of the conformers in the test set at the optimal value of $\alpha=0.5$ coloured by energy (a), radius of gyration (b), and peptide bond isomerism (c). Several individual conformers are also highlighted, including those with the highest and lowest energy and radius of gyration. For comparison, a KPCA projection of the same conformers is plotted in the inset in the upper right corner of each subplot. The KPCA projection alone represents well the different structural features (peptide bond isomerism and radius of gyration), but leads to rather poor energy regression ($\ell_\text{regr}=0.60$ as opposed to $\ell_\text{regr}=0.004$ for  KPCovR). 
The KPCovR projection separates more clearly a group of high-energy conformers, to the left, and a cluster of very stable configurations, to the lower right. 
The former are characterized by having both peptide bonds in the \emph{cis} configuration, and by an unfavourable steric interaction between the terminating methyl groups. 
The stable conformers, on the other hand, all have the naturally-preferred all-trans isomerism, and the backbone takes an extended $\beta$-strand structure. They only differ by the hydrogen-bonding pattern of the side-chain, that modulates in a more subtle way the conformational stability.  

\begin{figure*}
    \centering
    \includegraphics[width=\linewidth]{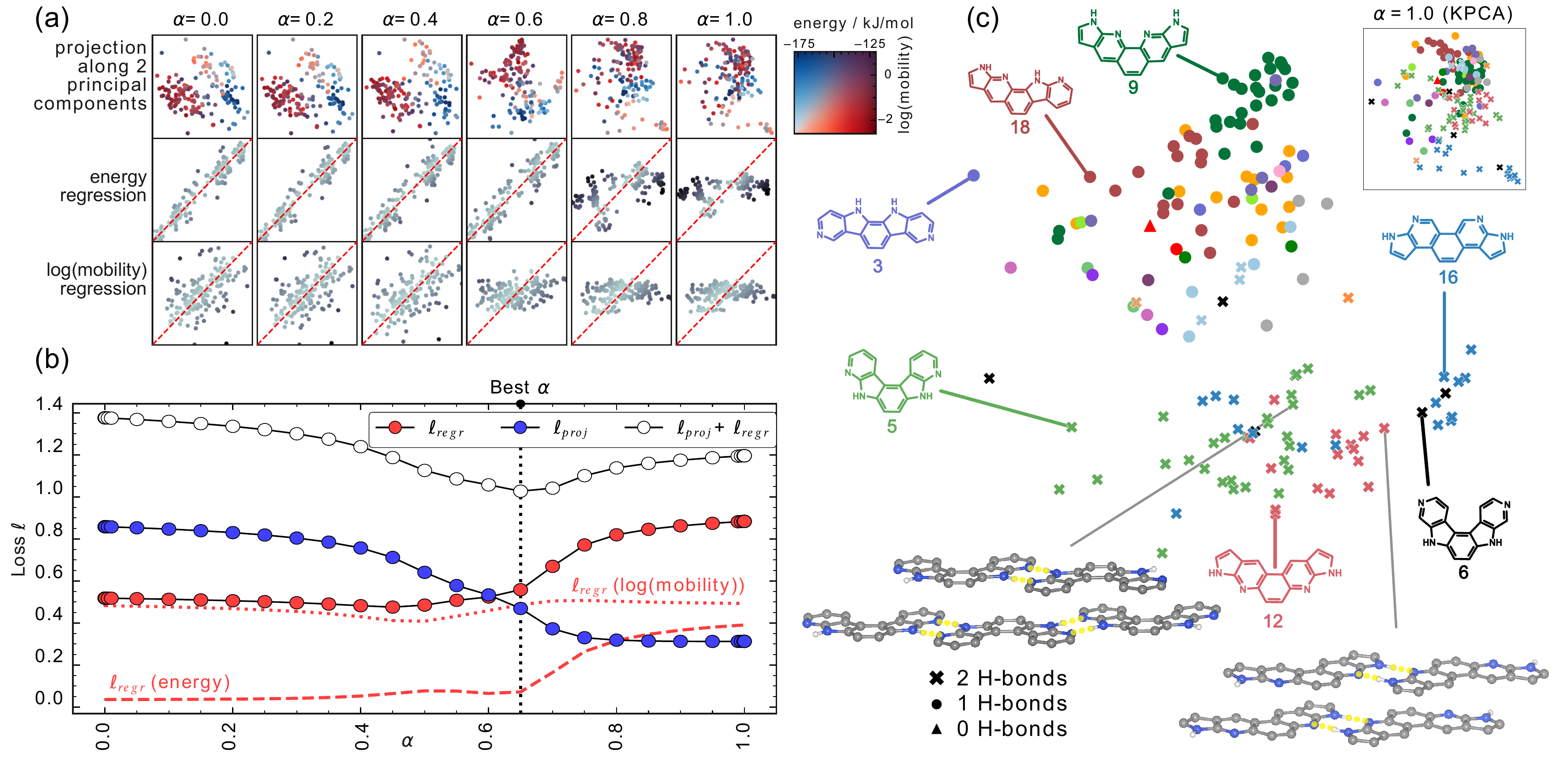}
    \caption{\textbf{KPCovR of Azaphenacenes.} (a) Overview of the projections produced by KPCovR for different $\alpha$ values coloured by both energy and electronic mobility according to the colour map on the side; parity plots for the regression of both energy and log of the electronic mobility over the testing set of Azaphenacenes (b) Evolution of the different losses as a function of alpha $\ell_\text{regr}$ is the total regression loss and KPCovR projection loss $\ell_\text{proj}$ (c) Annotated map of azaphenacenes as produced by KPCovR at
    $\alpha=0.65$. Each point corresponds to a different molecular crystal, with the colours and coloured insets denoting the substituent molecule.  Symbols show the number of hydrogen
    bond per molecule, as computed with PAMM\cite{Gasparotto2014}. On the bottom, we show two
    examples of DNA-style hydrogen bonds between molecules, the images were
    created using VESTA\cite{Momma2011}. The inset in the upper right contains the analogous KPCA projection.}
    \label{fig:azaphenacenes-overview}
\end{figure*}

The inclusion of an explicit supervised learning component in KPCovR does not only lead to a dimensionality reduction that preserves with much higher accuracy the underlying structure-property relations, but reveals more clearly the molecular motifs that stabilize (or de-stabilize) the different conformers.

\subsection{Azaphenacenes}

As a last example, we consider a dataset containing different crystallyne polymorphs of 28 isomers of a pyrrole-based azaphenacene compound, for which total energy and electronic mobility have been previously computed in Ref.~\citenum{Yang2018}.
This dataset present a series of challenges for KPCovR in particular and machine learning techniques in general, and as such is a good test for the new method presented in this paper.
First, the dataset contains a very small number of structures (311), half of which were used for training the different models.
Second, electronic mobility is an inherently non-local property, which makes it hard to predict it using local descriptors such as SOAP, even when the descriptors are grouped together in a \emph{structure} kernel as discussed in appendix~\ref{app:local-global}. It provides, therefore, a demonstration of the robustness of KPCovR in presence of target properties that are noisy, or otherwise un-learnable. 

We present the results of KPCovR on the validation set for different $\alpha$
values Fig.~\ref{fig:azaphenacenes-overview}, panels (a) and (b). On the pure regression, $\alpha = 0$ side, the prediction of energies is surprisingly good given the very low number of training points. The RMSE of 3.83 kJ/mol (equivalent to $\ell_\text{regr} = 0.038$) is around a quarter of the dataset in intrinsic standard deviation of 14.3 kJ/mol.
The prediction of electronic mobility is much harder. We used the logarithm of electronic mobilities instead of the raw values as the prediction target, as detailed in the supplementary information.
Although this transformation improved our ability to learn electronic mobilities, the regression loss is very high (RMSE of 0.9;  $\ell_\text{regr} = 0.482$ which is more than 90\% of the expected variance of 0.5). We are overall unable to learn electronic mobility for this dataset, similarly to what was already observed for this dataset\cite{Yang2018,musi+18cs}.

Looking now at the optimal $\alpha=0.65$ (i.e. the value of $\alpha$ minimising
the sum of the projection and regression losses), we observe that we are still unable to learn electronic mobility with a relative loss of 0.485 (equivalent to an error that is approximately 95\%{} of the intrinsic variability of the data, and only marginally worse of the error for $\alpha=0$), and the resulting prediction is visibly skewed in the parity plot. The poor regression performance for the log-mobility is reflected in the lack of a clear correlation between the position in latent space and the value of mobility.

Even if we are unable to predict the electronic mobility, the cohesive energy of the different polymorphs can be learned very effectively, and the optimal $\alpha=0.65$ corresponds to an excellent balance between $\ell_\text{regr}$ and $\ell_\text{proj}$. 
The latent-space projection separates the data set  in two clusters along the vertical axis. This separation is related to the a bi-modal energy distribution, with low and high energy structures, which is lost in the limit of pure KPCA at $\alpha=1$. In figure~\ref{fig:azaphenacenes-overview}, we show an annotated map of the different crystal stackings, coloured by molecular identity. 
Different symbols indicate the average number of hydrogen bonds between molecules in the crystals, which we identified using a Probabilistic Analysis of Molecular Motifs (PAMM)\cite{Gasparotto2014}. We find that the cluster of low energy stacking contains only structures with 2 hydrogen bonds per molecules (the maximal possible value), while the high energy cluster contains mostly structures with 1 hydrogen bond per molecule. Additionally, the majority of structures in the low energy cluster are linked by hydrogen bonds in a DNA-like fashion, i.e. by having pairs of matching hydrogen bond donors and acceptors facing each another. Finally, the low energy cluster only contains crystal created from molecules 5, 6, 12 and 16 (following the notation from the original paper\cite{Yang2018}), the 24 other molecules being in the high energy cluster. These four molecules are the only ones with just the right geometry to create matching, DNA-like hydrogen bonding patterns with two bonds per molecules, and high symmetry crystals.
Isomer 9 and 18 also contain a similar \ce{N-C-NH} motif, but cannot form a paired-HB pattern because of steric hindrance (9) and asymmetry(18). 

\section{Conclusions}

In this paper we provide a comprehensive overview of linear and kernel-based methods for supervised and unsupervised learning, showing an example of their application to elucidate and predict structure-property relations in solid-state NMR.
We also discuss a simple combination of principal component analysis and linear regression, PCovR~\cite{de_jong_principal_1992}, that has as yet received far less attention than in our opinion it deserves.
We derive extensions to PCovR that make it possible to use it in the context of kernel methods (KPCovR and sparse KPCovR), and demonstrate their application to five distinct datasets of molecules and materials.
We also prepared a set of Jupyter notebooks~\cite{kpcovr-notebooks} that provide a pedagogic introduction to both traditional and novel methods we discuss, and allow exporting structure-property maps in a format that can be visualised with an interactive tool that we also developed as part of this work\cite{chemiscope}.

The flexibility afforded by a kernel method allows improving substantially (typically by a factor of two) the regression performance relative to linear PCovR. Compared to kernel PCA, KPCovR maps reflect more explicitly structure-property relations, and -- in all the diverse cases we considered -- are more revealing, helping to identify the molecular motifs that determine the performance of the different structures, and that often reflect intuitive chemical concepts such as hybridisation, chemical composition, H-bond patterns. This study highlights the promise of combining supervised and unsupervised schemes in the analysis of data generated by atomistic modelling, to obtain clearer insights to guide the design of molecules and materials with improved performance, and to build more effective models to directly predict atomic-scale behaviour.

\begin{acknowledgments}
MC, RKC, BAH acknowledge funding by the from the European Research Council (Horizon 2020 grant agreement no. 677013-HBMAP). 
GF acknowledge support by the SCCER Efficiency of Industrial Processes, and by the European Center of Excellence MaX, Materials at the Hexascale - GA No. 676598.
\end{acknowledgments}

\section*{Supporting Information}

The electronic supporting information contains a comprehensive discussions of the parameters used for the analysis of each of the five examples, together with a comprehensive comparison of the performance of different linear, kernel and PCovR-like methods for the five data sets. 
For each data set we also provide an interactive map that can be visualized with the on-line viewer chemiscope~\cite{chemiscope}. 
A set of Juypyter notebooks that provide a hands-on tutorial for the application of KPCovR is available in a separate repository~\cite{kpcovr-notebooks}.

\appendix

\section{Centring and scaling}\label{app:centring}

In this paper, as it is often done in machine-learning applications, we centre and scale (or standardise) the original input data, which removes the dependency of results on a trivial shifting or scaling of the data set. centring and scaling are of particular importance in PCovR-based methods, as the model can be inherently biased towards the projection or regression if $\bX$ and $\bY$ data are of different relative magnitudes. To avoid ambiguity, we centre and scale our raw data $\bX^{\prime}$ and $\bY^{\prime}$ in the following manner,
\begin{align}
    \bX &= \frac{\sqrt{\ntrain}\left(\bXprime - \bXbarprime_{\text{train}}\right)}{\lVert\bXprime_{\text{train}} - \bXbarprime_{\text{train}}\rVert} \\
    \bY_i &= \frac{\sqrt{\ntrain}\left(\bYprime_i - \bYbarprime_{i, \text{train}}\right)}{\sqrt{\np}\lVert\bYprime_{i, \text{train}} - \bYbarprime_{i, \text{train}}\rVert},
\end{align}
where $\bA_i$ denotes the $i^{th}$ property (column) of $\bA$, $\bar{\bA}$ is the columnwise mean of $\bA$, and $\bA_{\text{train}}$ indicates the subset of samples in $\bA$ that belong to the training set. By centring and scaling the data in this manner, we ensure that the squared Frobenius norms of $\bX_{\text{train}}$ and $\bY_{\text{train}}$ are equal to $\ns$, and that individual property variances of $\bY_{\text{train}}$ are all equal to $1/\np$.

We perform a similar centring and scaling procedure when constructing kernels. Kernel standardization can be viewed as simply centring and scaling the data in the RKHS feature space. If $N$ indicates the dataset that defines the centring (typically the train set) and $i$, $j$ two data points between which we want to compute the centred kernel,
\begin{equation}
K_{ij}=\frac{n_N\left(\bphi_i-\bar{\bPhi}_N\right)^T\left(\bphi_j-\bar{\bPhi}_N\right)}{\Tr \left(\left(\bPhi_N-\bar{\bPhi}_N\right)\left(\bPhi_N-\bar{\bPhi}_N\right)^T\right)}
\end{equation}
where $\bar{\bPhi}_N$ is the column mean of the training set feature matrix $\bPhi_N$, and is computed once and for all for the train set, together with the normalisation factor.
This can be written avoiding to compute explicitly the RKHS features:
\begin{equation}
K_{ij}=\frac{n_\text{train}}{\Tr \bG_{NN}}\left(
K'_{ij}-\sum_{n\in N} \frac{K'_{in}+K'_{jn}}{n_\text{train}} + \sum_{nn'\in N}\frac{K'_{nn'}}{n_\text{train}^2}\right).
\end{equation}
Centring is achieved by computing column averages of the raw kernels between points $i$, $j$ and the train set points. Note that kernel matrix elements may refer to different matrices, depending on whether $i$ and $j$ are themselves train set points, or new inputs.
The scaling factor $n_\text{train}/\Tr \bG_{NN}$ is computed using the \emph{centred} train set kernel.

In sparse KPCA and sparse KPCovR, a slightly different approach is required, as the goal is to ensure that the Nystr\"{o}m approximation to the full kernel matrix is centred and scaled properly -- i.e. the active set kernel defines the RKHS, but centring and scaling should be computed based on the training set $N$. A centred and scaled kernel between an input $i$ and an active point $m\in M$ can then be computed as
\begin{equation}
K_{im}=\frac{n_\text{train}}{\sqrt{\Tr\left(\bG_{NM}\bG_{MM}^{-1}\bG_{NM}^T\right)}}\left(
K'_{im}-\sum_{j\in N} \frac{K'_{jm}}{n_\text{train}}\right),
\end{equation}
where once more the normalisation factor is computed using the centred version of $\bG_{NM}$.

\section{Projection loss in kernel methods and KPCovR}
\label{app:kpcovr-loss}
Rewriting Eq.~\eqref{eq:loss-pcovr} in terms of the RKHS, we get:
\begin{equation}
    \ell = \alpha\lprojphi + (1 - \alpha)\lregr.
\end{equation}
The latter portion of this equation, $\ell_\text{regr}$, can be written in terms of the KRR loss given in Eq.~\eqref{eq:krr-loss}, where $\bPpy$ also encapsulates the loss incurred from the latent space projection.
The former portion, $\ell_\text{proj}$ is straightforward to compute given an explicit RKHS. In case one wants to avoid evaluating the RKHS, however, $\ell_\text{proj}$ may be computed in terms of the kernel.

Indicating the kernel between set $A$ and $B$ as $\bG_{AB}$, the projection of set $A$ as $\bT_A$, and with N and V as the train and validation/test set, one obtains
\begin{equation}
\begin{split}
\ell_\text{proj}=&
\Tr\left[
\bG_{VV} - 2
    \bG_{VN} \bT_N (\bT_N^T \bT_N)^{-1}  \bT_V^T\right.\\
    +&\bT_V(\bT_N^T \bT_N)^{-1}  \bT_N^T   \bG_{NN} \left.\bT_N (\bT_N^T \bT_N)^{-1}    \bT_V^T .
\right]
\end{split}
\end{equation}
When the loss is evaluated on the train set, so that $N\equiv V$, this expression reduces to
\begin{equation}
      \ell_\text{proj} = \Tr\left(\bG_{NN} - \bG_{NN} \bPkt \bPtk \right).
\end{equation}
where $\bPtk = (\bT_N^T \bT_N)^{-1} \bT_N^T \bG_{NN}$.
A full derivation of this loss equation can be found in the SI.

\section{Structures and environments}
\label{app:local-global}

When analyzing molecular or materials structures, there are several possible scenarios, involving the prediction of atom-centred or global properties, and the search for structural correlations between atomic environments or overall structures.  Whenever the nature of the property and that of the structural entity match, the formalism we have reviewed in Section~\ref{sec:methods} applies straightforwardly.
A common scenario that deserves a separate discussion involves the case in which one seeks to reveal how atomic environments or molecular fragments contribute to global properties of a material. Often, this means that the properties of a structure $\by(\CA)$ are written as a sum over contributions from the atom-centred environments in each structure, $\by(\CA)=\sum_{i\in\CA} \by(\CX_i)$.
For a linear model, this means that the regression loss reads
\begin{equation}
\begin{split}
\ell_\text{regr}=&\frac{1}{\ns}\sum_{\CA} \loss{\by(\CA)}{\sum_{i\in\CA} \bx_{i} \bP{X}{Y}} = \\ =&\frac{1}{\ns} \sum_{\CA} \loss{\by(\CA)}{\tilde{\bx}(\CA) \bP{X}{Y}},
\end{split}
\end{equation}
where we defined $\tilde{\bx}(\CA)=\sum_{i\in\CA} \bx_{i}$.
In other terms, one can formulate the regression using features that describe the structures as a sum of the features of their atoms, and then proceed to determine the weights $\bP{X}{Y}$ as in conventional linear regression. A similar expression holds for kernel  methods, where the kernels between structures can be built as sums over kernels between environments, resulting in an additive property model.
In a (K)PCovR framework, where one is restricted to learning the fraction of the properties that can be approximated as a (kernelized) linear function of $\bx$, one should first train a model based on the full structures, and then compute the predictions for individual environments. These are combined to form the approximate property matrix $\bYhat$. The model can then be built as in the homogeneous case of environment features and atom-centred properties.
%

%

%
%
%
%

%
%
%
%

\end{document}